\documentclass[10pt,twocolumn,letterpaper]{article}

\usepackage{cvpr}              
\usepackage{tabularx}
\usepackage{graphicx}
\usepackage{tabularray}
\usepackage{threeparttable}
\usepackage{colortbl}
\usepackage{xcolor}
\usepackage{pgfplotstable} 
\usepackage{colortbl}      
\definecolor{mygreen}{RGB}{59, 95, 33}   
\definecolor{myorange}{RGB}{239, 134, 53}   
\definecolor{mypurple}{RGB}{112, 48, 160}   
\usepackage{multirow}
\usepackage{amssymb}
\usepackage{pifont}
\definecolor{cvprblue}{rgb}{0.21,0.49,0.74}
\usepackage[pagebackref,breaklinks,colorlinks,allcolors=cvprblue,linkcolor=red]{hyperref}
\usepackage[accsupp]{axessibility}  

\def\paperID{12876} 
\def\confName{CVPR}
\def\confYear{2025}

\title{DroneSplat: 3D Gaussian Splatting for Robust 3D Reconstruction from In-the-Wild Drone Imagery}

\author{Jiadong Tang\texorpdfstring{\textsuperscript{1}}{1} \quad 
Yu Gao\texorpdfstring{\textsuperscript{1}}{1} \quad 
Dianyi Yang\texorpdfstring{\textsuperscript{1}}{1} \quad 
Liqi Yan\texorpdfstring{\textsuperscript{2}}{2} \quad 
Yufeng Yue\texorpdfstring{\textsuperscript{1}}{1}\quad 
Yi Yang\textsuperscript{1}\textsuperscript{,}\footnotemark[1]\\
\textsuperscript{1}{Beijing Institute of Technology}  \quad
\textsuperscript{2}{Hangzhou Dianzi University} \\
}

\begin{document}

\twocolumn[{
\renewcommand\twocolumn[1][]{#1}
\maketitle
\begin{center}
    \centering
    \includegraphics[width=\textwidth]{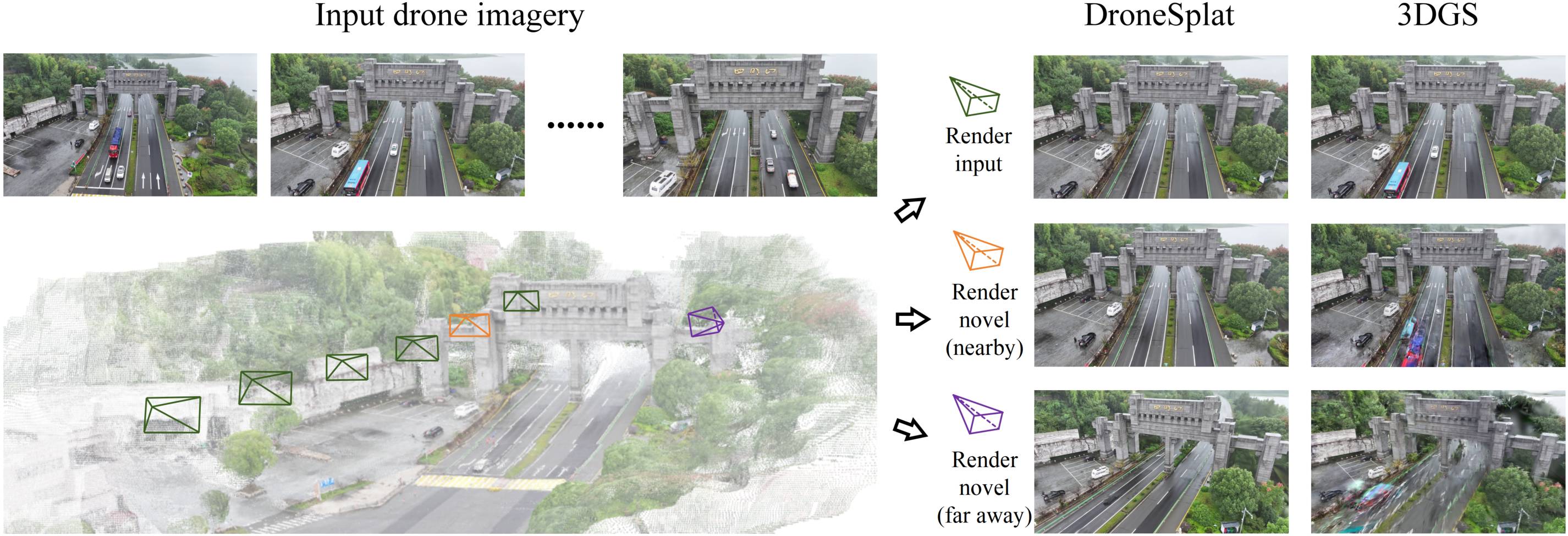}
    \captionof{figure}{Given a set of drone imagery, our method effectively eliminates the impact of dynamic distractors on the static scenes (e.g., vehicles driving on the road). The right side of the figure shows the rendering results of our method compared to 3DGS under \textcolor{mygreen}{input view} and \textcolor{myorange}{novel view}. Our method eliminates distractors while 3DGS generates artifacts in the corresponding regions. Moreover, our method reconstructs scenes with accurate geometry under limited viewpoints, demonstrating robustness to significant view variation (\textcolor{mypurple}{novel view}).}
    \label{img1}
\end{center}%
}]
\renewcommand{\thefootnote}{\fnsymbol{footnote}} 
\footnotetext[1]{Corresponding author: Yi Yang (yang\_yi@bit.edu.cn).} 
\begin{abstract}

Drones have become essential tools for reconstructing wild scenes due to their outstanding maneuverability. Recent advances in radiance field methods have achieved remarkable rendering quality, providing a new avenue for 3D reconstruction from drone imagery. 
However, dynamic distractors in wild environments challenge the static scene assumption in radiance fields, while limited view constraints hinder the accurate capture of underlying scene geometry.
To address these challenges, we introduce DroneSplat, a novel framework designed for robust 3D reconstruction from in-the-wild drone imagery.
Our method adaptively adjusts masking thresholds by integrating local-global segmentation heuristics with statistical approaches, enabling precise identification and elimination of dynamic distractors in static scenes.
We enhance 3D Gaussian Splatting with multi-view stereo predictions and a voxel-guided optimization strategy, supporting high-quality rendering under limited view constraints.
For comprehensive evaluation, we provide a drone-captured 3D reconstruction dataset encompassing both dynamic and static scenes.
Extensive experiments demonstrate that DroneSplat outperforms both 3DGS and NeRF baselines in handling in-the-wild drone imagery.
Project page: \href{https://bityia.github.io/DroneSplat/}{\texttt{https://bityia.github.io/DroneSplat/}}.
\end{abstract}    
\vspace{-1em}
\section{Introduction}
3D reconstruction is pivotal in fields such as cultural heritage preservation \cite{GOMES20143}, geological investigation \cite{drones7030198} and urban surveying \cite{liu2025citygaussian, turki2022mega}. In this context, drones have become valuable for these tasks due to their outstanding maneuverability. 
Capable of traversing obstacles like water and difficult terrain, drones enable extensive data acquisition from varied altitudes and angles.

\begin{figure}[!t]
    \centering
    \includegraphics[width=\linewidth]{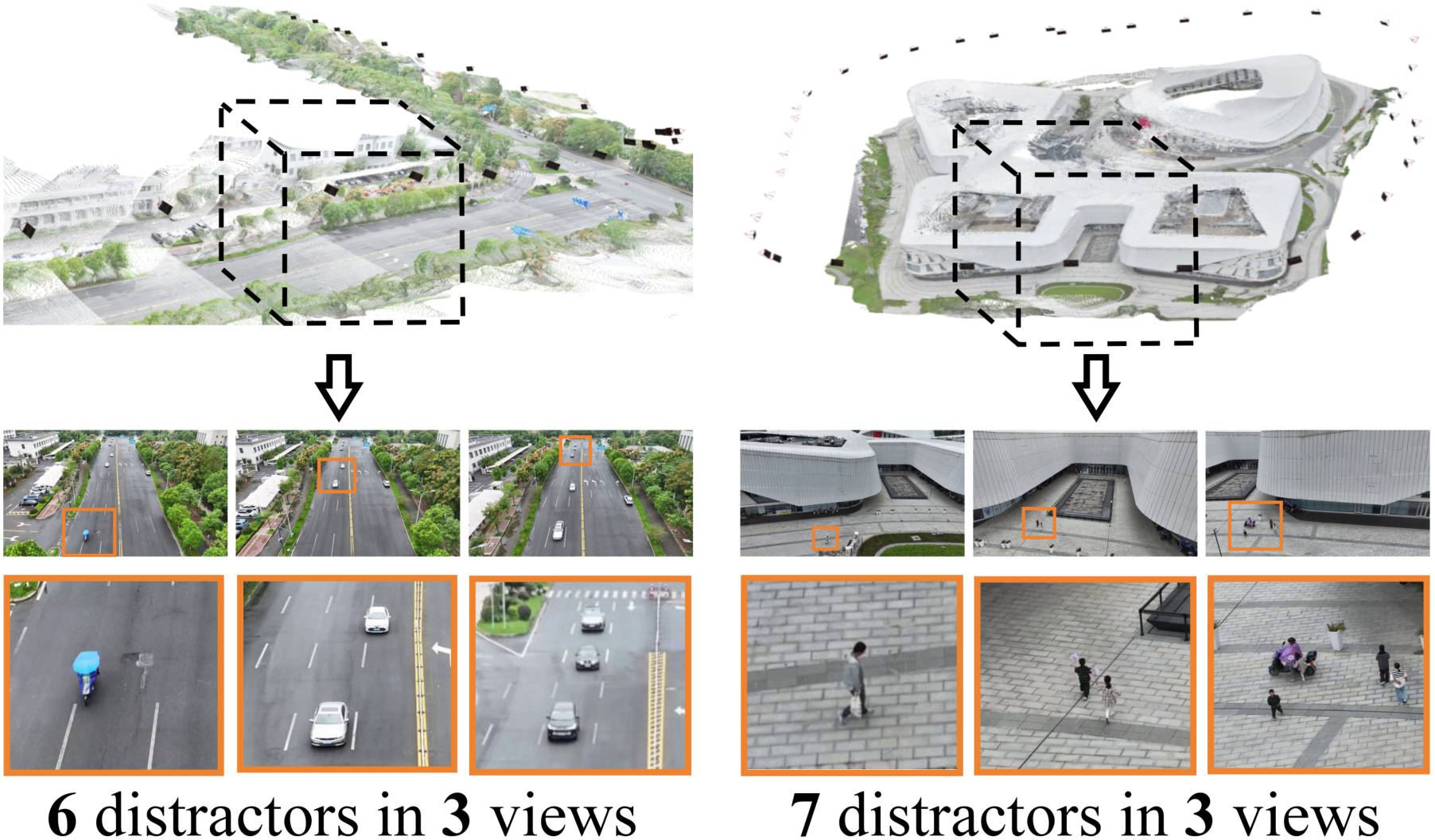}
    \caption{\textbf{The challenges within in-the-wild drone imagery.} Certain regions within the entire scene will face the challenge of dynamic distractors and limited view constraints. }
    \label{challenges}
     \vspace{-1em} 
\end{figure}

Recently, radiance field methods, such as NeRF \cite{mildenhall2020nerf} and 3D Gaussian Splatting (3DGS) \cite{kerbl3Dgaussians}, have shown remarkable potential in 3D representation and novel view synthesis. Compared to traditional drone-based 3D reconstruction methods like oblique photogrammetry, NeRF and 3DGS can capture more realistic surface while reducing the training time. However, applying NeRF or 3DGS to in-the-wild drone imagery presents several challenges for high-quality 3D reconstruction (Figure \ref{challenges}).

The primary challenge is \textit{scene dynamics}. 
Drone-captured images in wild scenes often contain moving objects, or distractors, which violate the assumption of multi-view consistency in NeRF or 3DGS.
To eliminate the impact of distractors, some works \cite{rematas2022urban, sun2022neural, tancik2022block, turki2022mega} attempt to identify and mask specific moving objects by pre-trained semantic segmentation models. However, these methods rely on predefined categories of potential distractors and fail to distinguish static and dynamic objects of the same category, limiting their effectiveness in uncontrolled wild scenes.
Other methods \cite{martin2021nerf, chen2022hallucinated, kim2024up, li2023nerf, ren2024nerf, zhang2024gaussian, kulhanek2024wildgaussians, sabour2023robustnerf,chen2024nerf} utilize reconstruction residuals to predict distractors, which work well on controlled datasets, but struggle with smaller distractors in drone imagery, making accurate pixel-level mask prediction challenging.
Additionally, existing approaches tend to apply hard thresholding to distinguish dynamic distractors, overlooking its limitations in adapting to variations across different scenes and training stages.
The second challenge is \textit{viewpoint sparsity}.
In a single flight, drone imagery oftens lacks comprehensive coverage of all viewpoints.
For example, when flying over a street, the entire scene may include numerous images, but the viewpoints of certain regions remain limited (the black dashed box in Figure \ref{challenges} is covered by only three views). Similar to challenges faced in sparse-view 3D reconstruction \cite{zhu2025fsgs, li2024dngaussian}, radiance fields may overfit to the inputs in limited view constraints, leading to poor rendering quality for novel views. Previous approaches attempt to address this by incorporating additional priors \cite{wang2023sparsenerf, deng2022depth, zhou2023sparsefusion, wu2024reconfusion, wynn2023diffusionerf} or designing hand-crafted heuristics \cite{niemeyer2022regnerf, yang2023freenerf, yu2021pixelnerf}. More recently, InstantSplat \cite{fan2024instantsplat} propose a novel paradigm that  initializes Gaussian primitives from multi-view stereo predictions to overcome sparsity. However, despite incorporating geometric priors, InstantSplat lacks corresponding optimization in 3DGS, undermining the abundant priors.

To address these challenges, we introduce DroneSplat, a robust 3D gaussian splatting framework tailored for in-the-wild drone imagery. 
For the issue of \textit{scene dynamics}, 
we propose Adaptive Local-Global Masking, which integrates the strengths of segmentation heuristics and statistical approaches to enhance distractor identification and elimination. It can autonomously adjust the local masking threshold based on real-time residuals and pixel-level segmentation results, while tracking high residual candidates within the scene context as global masking. 
For the issue of \textit{viewpoint sparsity},
our framework employs a multi-view stereo model to provide rich geometric priors by predicting dense 3D points. Based on these points, we implement a geometric-aware point sampling alongside a voxel-guided optimization strategy to effectively constrain 3DGS optimization.

Furthermore, to rigorously evaluate our approach in wild scenes, we provide a dataset of 24 drone-captured sequences, encompassing both dynamic and static scenes. The dynamic scenes are further categorized into three levels based on the number of dynamic distractors. 

Our contributions are the following:
\begin{itemize}

\item We introduce a robust 3D Gaussian splatting framework tailored for in-the-wild drone imagery, which eliminates the impact of dynamic distractors on static scene reconstruction and addresses the issue of unconstrained optimization in 3DGS under limited viewpoints.

\item We propose Adaptive Local-Global Masking that harmoniously integrates local-global segmentation heuristics and statistical approaches,  combining local and global masking as collaborators to deliver robust and superior performance in dynamic distractor elimination.

\item We enhance 3DGS with a geometric-aware point sampling method and a voxel-guided optimization strategy, effectively leveraging multi-view stereo’s rich priors alongside 3DGS’s powerful representational capacity.

\item We provide a drone-captured 3D reconstruction dataset of 24 in-the-wild sequences, encompassing both dynamic and static scenes.

\end{itemize}
\section{Related Work}
\textbf{Drone based 3D reconstruction.}
A typical method for 3D reconstruction with drones is oblique photography, where drones equipped with multiple cameras capture images from elevated angles. 
In traditional oblique photography, camera poses are estimated using a Structure-from-Motion (SfM) pipeline, followed by surface reconstruction through dense multi-view stereo \cite{zeng2024oblique}. Recently, radiance field methods \cite{mildenhall2020nerf, kerbl3Dgaussians} have achieved realistic renderings. Several works \cite{jia2024drone, maxey2024uav, turki2022mega, zeng2024oblique} utilize NeRF \cite{mildenhall2020nerf} to reconstruct scene from drone-captured images, and most prioritize the challenges associated with large-scale scene reconstruction. DRAGON \cite{ham2024dragon} optimizes a 3D Gaussian model \cite{kerbl3Dgaussians} with drone imagery and ground building images. While previous work in radiance fields have advanced static scene reconstruction, these approaches encounter substantial challenges in Non-static environments.

\begin{figure*}[!h]
    \centering
    \includegraphics[width=\textwidth]{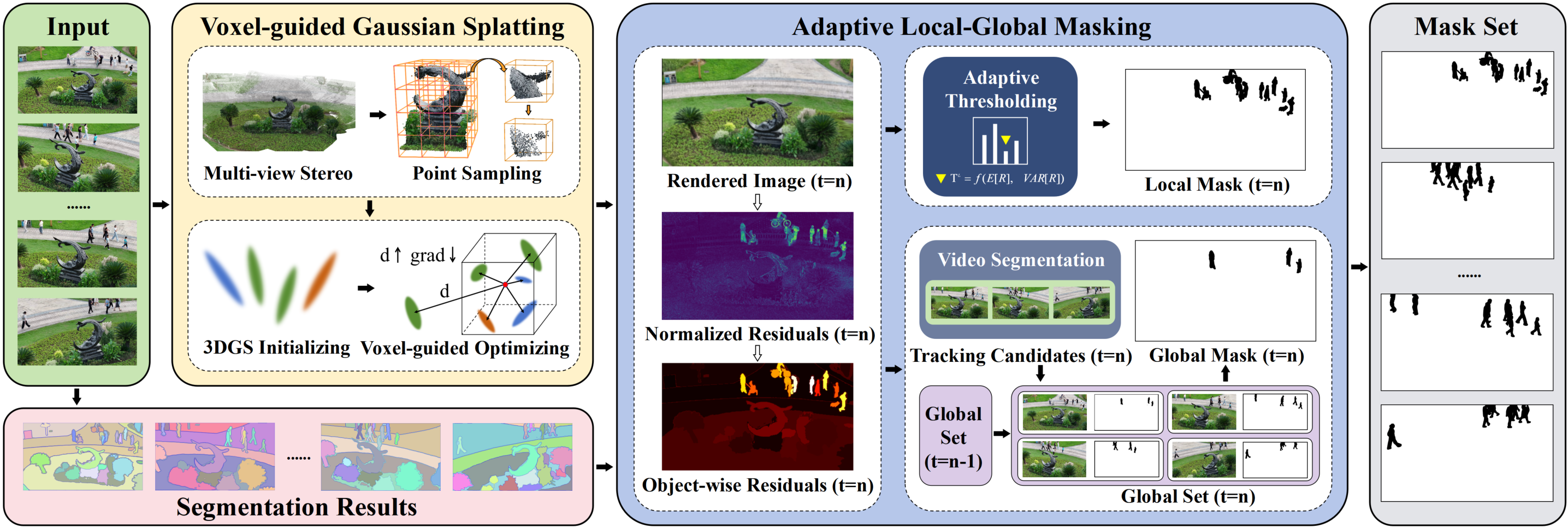}
    \caption{\textbf{The framework of DroneSplat.} Given a few posed drone imagery of a wild scene, our goal is to identify and eliminate dynamic distractors. We first predict a dense point cloud through a learning-based multi-view stereo method, followed by point sampling based on confidence and geometric features. The sampled point cloud is used to initialize Gaussian primitives, which is then optimized using a voxel-guided strategy. At iteration $t=n$, we calculate the normalized residual of the rendered image and combine it with the segmentation results to obtain the object-wise residuals. We adaptively adjust the threshold based on the object-wise residuals and statistical approaches to obtain local masks. Meanwhile, we mark objects with high residuals as tracking candidates, deriving the global set at $t=n$ by combining the global set at $t=n-1$ with the tracking outcomes at $t=n$. After merging the local mask and the global mask retrieved from the global set, we can obtain the final mask at time $ t= n$. The mask set illustrates the dynamic distractors we predicted.}
    \label{pipeline}
    \vspace{-1em}
\end{figure*}

\noindent\textbf{Radiance fields in Non-static Scenes.}
There are roughly two approaches for reconstructing dynamic scenes: one aims to reconstruct the complete process of the scene changes from video sequences \cite{huang2024sc, li2024spacetime, yu2024cogs}, while the other attempts to eliminate the influence of dynamic distractors to achieve static scene reconstruction. Our work aligns with the latter. Previous works in this realm have focused on how to accurately identify dynamic objects, or distractors, which can be roughly classified into two paradigms:
\begin{itemize}

\item Segmentation-based methods \cite{rematas2022urban, sun2022neural, tancik2022block, turki2022mega, otonari2024entity, chen2024nerf} rely on pretrained semantic segmentation model to mask common dynamic objects like pedestrians. They require predefined classes for distractors, limiting their adaptability in real-world scenarios. Moreover, solely depending on semantic segmentation may lead to misclassification static objects as dynamic ones, such as parked cars along the roadside. Recent work NeRF-HuGS \cite{chen2024nerf} enhances the traditional heuristic design by integrating the strengths of structure-from-motion and color residuals to obtain prompt points for SAM \cite{kirillov2023segment}, thereby improving dynamic distractor segmentation. However, this approach requires manually tuning thresholds for each scene, and these threshold remain fixed during training, overlooking the residual variance at different training stages.
 
\item Uncertainty-based methods \cite{martin2021nerf, chen2022hallucinated, kim2024up, li2023nerf, ren2024nerf, zhang2024gaussian, kulhanek2024wildgaussians, sabour2023robustnerf} employ color residuals and image feature to predict uncertainty (distractors), making them more adaptable in uncontrolled environments as they do not rely on priors of the scene. NeRF On-the-go \cite{ren2024nerf} and WildGaussians \cite{kulhanek2024wildgaussians} utilize DINOv2 \cite{oquab2023dinov2} to extract features and design a network to predict the uncertainty. However, predicting pixel-level masks from low-resolution features results in blurred edge details and omission of small objects, posing challenges for accurately identification, especially in drone imagery where the distractors tend to be small. 

\end{itemize}

\noindent\textbf{Radiance fields under limited views.}
The performance of radiance fields substantially degrades when the input views become sparse. NeRF-based apporaches focus on incorporate regularization constraints \cite{yang2023freenerf, niemeyer2022regnerf, yu2021pixelnerf} or introduce extra priors, such as depth estimation model \cite{wang2023sparsenerf, deng2022depth} and diffusion model \cite{zhou2023sparsefusion, wu2024reconfusion, wynn2023diffusionerf}. 3DGS-based work \cite{zhu2025fsgs,li2024dngaussian,chung2024depth, xiong2023sparsegs} typically integrate depth priors with hand-crafted heuristics. 
Recently, InstantSplat \cite{fan2024instantsplat} has demonstrated notable achievement by leveraging the dense points predicted by DUSt3R \cite{wang2024dust3r} as geometric priors to initialize Gaussian primitives. However, Instantsplat lacks constraints for subsequent Gaussian optimization, leading to suboptimal performance despite the presence of rich priors. 
\section{Method}

\subsection{Preliminaries}
We build our work upon 3DGS \cite{kerbl3Dgaussians}, which represents a 3D scene explicitly through a collection of 3D anisotropic Gaussian primitives. Specifically, a Gaussian primitive can be defined by the following set of parameters: a center point $\mu$, a scaling matrix $S$, a rotation matrix $R$, an opacity parameter $\alpha$ and a color feature $f$. The influence of each Gaussian on a 3D point $x$ can be described:
\begin{equation}
G(x)=e^{-\frac{1}{2}(x-\mu)^{T}\Sigma^{-1}(x-\mu)}
\end{equation}
where $\Sigma$ is a positive semi-definite covariance matrix which can be calculated from the scale matrix $S$ and the rotation matrix $R$. To render 2D image, 3DGS first project Gaussian primitives onto the 2D plane, and then blending $N$ ordered Gaussians overlapping each pixel to compute the color:
\begin{equation}
c=\sum\limits_{i=1}^{N} c_i\tilde{\alpha}_{i}\prod\limits_{j=1}^{i-1} (1-\tilde{\alpha}_{j})
\end{equation}
where $c_{i}$ is the decoded color of feature $f$, $\tilde{\alpha}_{i}$ represents the opacity of the projected 2D Gaussians.

3DGS is optimized by a combination of D-SSIM \cite{wang2004image} and L1 loss computed from the rendered color and the ground truth color:
\begin{equation}
\mathcal{L} = (1-\lambda_{dssim})\mathcal{L}_{\text{L1}} + \lambda_{dssim}\mathcal{L}_{\text{D-SSIM}} \label{loss}
\end{equation}
where $\lambda_{dssim}$ is a weighting factor.
\subsection{Adaptive Local-Global Masking}
The predominant approach for eliminating distractors in static scene reconstruction is to identify masks $\mathcal{M}$ for dynamic distractors and incorporate them to Eq. \ref{loss}. The training loss can be defined as follows:
\begin{equation}
\mathcal{L} = (1-\lambda_{dssim})\mathcal{M}\mathcal{L}_{\text{L1}} + \lambda_{dssim}\mathcal{M}\mathcal{L}_{\text{D-SSIM}} \label{train_loss}
\end{equation}

The training loss (Eq. \ref{train_loss}) indicates that higher mask accuracy leads to fewer artifacts and improved reconstruction quality.  

\noindent\textbf{Object-wise Average of Normalized Residuals.}
Inspired by RobustNeRF \cite{sabour2023robustnerf}, we start with the residuals between the rendered image and the ground truth.
Note that all residuals mentioned later are calculated independently and are not involved in the loss function.
We normalize the DSSIM residual $\mathcal{R}_{\text{D-SSIM}}$ and L1 residual $\mathcal{R}_{\text{L1}}$ to $\tilde{\mathcal{R}}_{\text{D-SSIM}}$ and $\tilde{\mathcal{R}}_{\text{L1}}$, respectively, mitigating the impact of residual variations that could affect distractor identification. Then we can obtain the combined residuals $\tilde{\mathcal{R}}$ as:
\begin{equation}
\tilde{\mathcal{R}} = (1-\lambda_{dssim})\tilde{\mathcal{R}}_{\text{L1}} + \lambda_{dssim}\tilde{\mathcal{R}}_{\text{D-SSIM}}
\end{equation}



To achieve pixel-level mask precision without reliance on predefined classes, we employ the latest Segment Anything Model v2 \cite{ravi2024sam}. For each image $I_{i}$ in the training set $I$,  we use the segmentation model $S$ to obtain $S(I_{i})= \{m_{i}^{1}, m_{i}^{2}, \cdots, m_{i}^{N_{i}}\}$, where $m_{i}^{j}$ represents the segmentation mask of the $j$-th object in image $I_{i}$ and $N_{i}$ denotes the number of masks of image $I_{i}$.

We assume that all pixels of the same object are treated jointly, sharing a unified residual value. The object-wise average residual $\mathcal{R}_{i}^{j}$ for a mask $ m_{i}^{j} $ can be calculated as:
\begin{equation}
\mathcal{R}_{i}^{j}(t) = \frac{\sum_{p \in m_{i}^{j}} \tilde{\mathcal{R}}(p,t)}{|m_{i}^{j}|}
\end{equation}

where $ p \in m_{i}^{j} $ indicates that $ p $ is one of the pixel positions contained within the mask $ m_{i}^{j} $, $t$ is the current iteration and $ |m_{i}^{j}| $ is the area (number of pixels) of mask $ m_{i}^{j} $.

\noindent\textbf{Adaptive Local Masking.}
Since dynamic objects are more difficult to fit and often exhibit higher residuals, a straightforward approach is to set a threshold to identify distractors. However, this process can be tedious: a higher threshold may fail to remove large dynamic objects, while a lower threshold risks misclassifying static objects as distractors \cite{chen2024nerf}. Previous studies \cite{sabour2023robustnerf, otonari2024entity, chen2024nerf} have not fully addressed this issue, instead applying a fixed threshold throughout training.

To establish an accurate and appropriate threshold across different scenarios and training stages, we propose an adaptive method to adjust threshold based on real-time residuals and statistical approaches. For the current training frame, we calculate the mathematical expectation and variance of the pixel-wise average residual:
\begin{equation}
\mathbb{E}[R_{i}](t) = \sum_{k=1}^{N} p_k \cdot r_k(t) 
\end{equation}
\begin{equation}
\text{Var}[R_{i}](t) = \frac{1}{N} \sum_{k=1}^{N} (r_k(t) - \mathbb{E}[R_{i}](t))^2
\end{equation}

where $N$ denotes the number of pixels in the current frame, $t$ is the current iteration, $  r_k(t) \in { R_i(t) } $ is residual of the $k$-th pixel, and $p_k$ represents the probability of the $k$-th pixel, typically $ p_k = \frac{1}{N} $ if all pixels are equally weighted. 

It can be observed that, across various scenes and training stages, nearly all static objects fall within one standard deviation of the expectation. In other words, pixels with residuals exceeding the expectation plus the variance can be regarded as distractors.
Considering the varying convergence rates of residuals across objects in the early stage of training, we relax the upper limit of the threshold. We set $\mathcal{T}_{i}^{L}$ as local threshold for image $I_{i}$:
\begin{equation}
\mathcal{T}_{i}^{L}(t) = \mathbb{E}[R_{i}](t) + \text{Var}[R_{i}](t)(1 + \lambda_{L}\frac{T_{max}-t}{T_{max}})
\end{equation}

where $T_{max}$ is the maximum iteration, $t$ is the current iteration and $\lambda_{L}$ is a weighting factor.

\begin{figure}[!t]
    \centering
    \includegraphics[width=\linewidth]{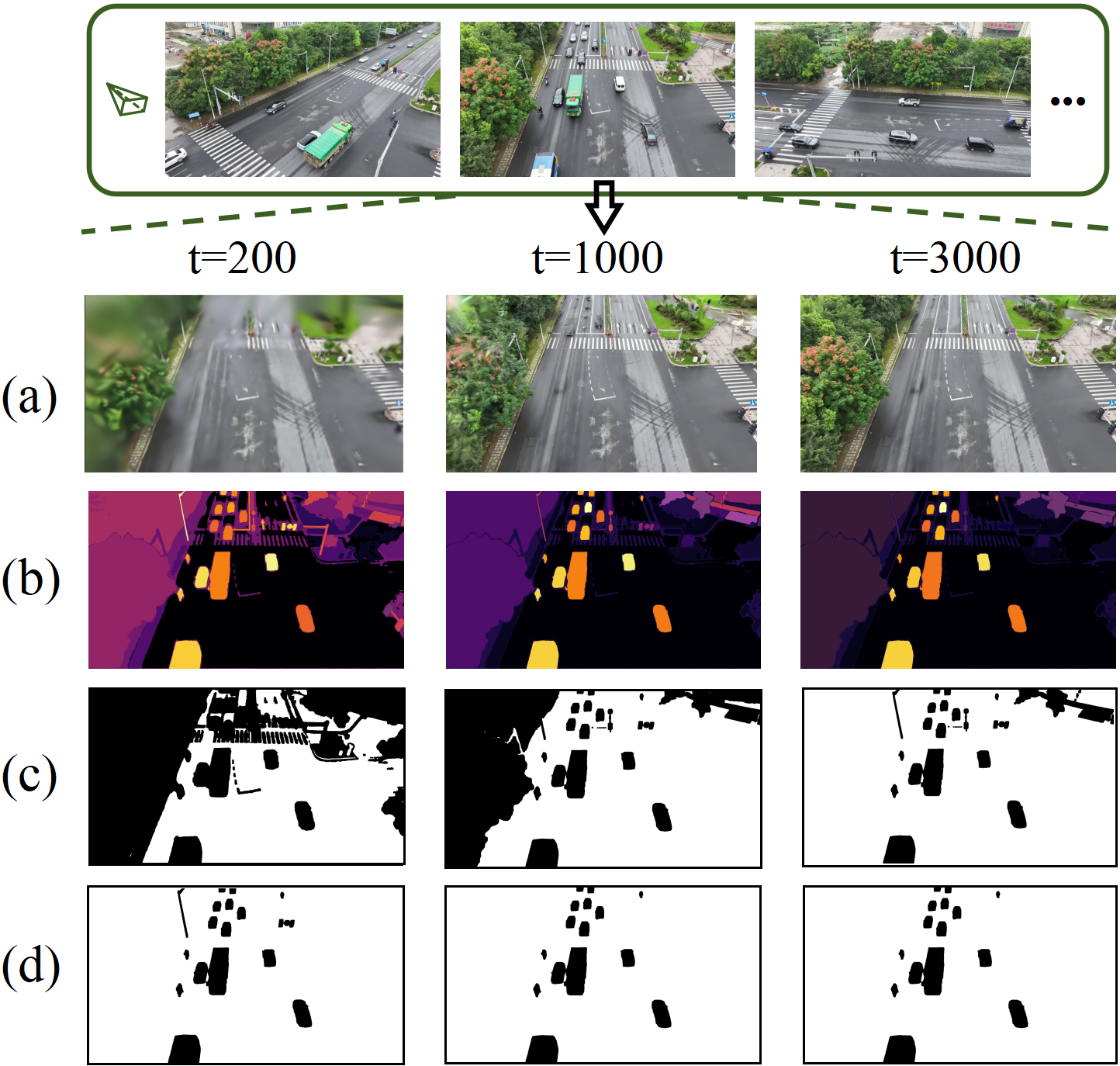}
    \caption{\textbf{The effect of Adaptive Local Masking.} (a) represent the renderings of the same frame across different iterations $t$, and (b) show the corresponding object-wise residuals. (c) are the masks obtained using a hard threshold, while (d) are the masks obtained by Adaptive Local Masking. }
    \label{local_mask}
     \vspace{-1em} 
\end{figure}

Let $ \mathcal{M}_{i}^{L}(t) $ denote the set of local masks $ m_{i}^{j} $ in image $ I_i $ at iteration $t$ whose object-wise average residual $ \mathcal{R}_{i}^{j}(t) $ exceeds the threshold $ \mathcal{T}_{i}^{\text{L}}(t) $:
\begin{equation}
\mathcal{M}_{i}^{L}(t) = \left\{ m_{i}^{j} \mid \mathcal{R}_{i}^{j}(t) > \mathcal{T}_{i}^{L}(t), \, j \in N_i \right\}
\end{equation}
where $ N_i $ is the total number of masks (objects) in image $ I_i $, and $ \mathcal{R}_{i}^{j}(t) $ represents the object-wise average residual for mask $ m_{i}^{j} $ at iteration $t$. Figure \ref{local_mask} illustrates the effect of Adaptive Local Masking.

\begin{figure}[!t]
    \centering
    \includegraphics[width=\linewidth]{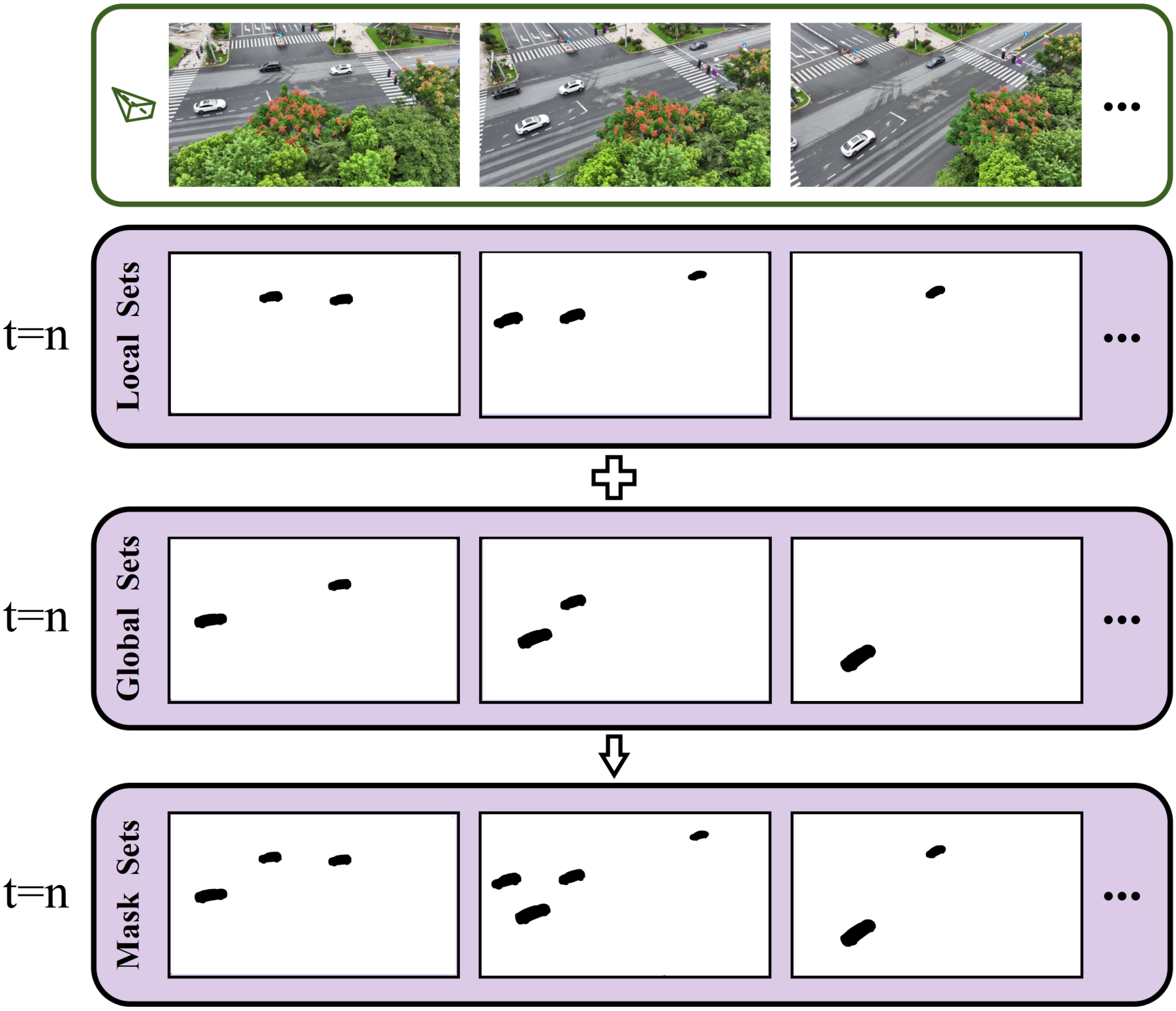}
    \caption{\textbf{The effect of Complement Global Masking.} At $t=n$, the white car waiting at a red light is not identified by the Adaptive Local Masking, but it is tracked through Complement Global Masking in other frames.}
    \label{global_mask}
     \vspace{-1em} 
\end{figure}

\noindent\textbf{Complement Global Masking.}
While Adaptive Local Masking is effective in identifying distractors in each frame, it still has certain limitations when handling some specific scenes. For example, an object may remain stationary in several frames but still should be considered a distractor in the overall context, such as a moving vehicle stops at a red light at an intersection (Figure \ref{global_mask}).

To address this, we employ the video segmentation capabilities of Segment Anything Model v2 \cite{ravi2024sam}. We set $\mathcal{T}_{i}^{G}$ as global threshold for image $I_{i}$:
\begin{equation}
\mathcal{T}_{i}^{G} = \mathbb{E}[R_{i}](t) + \lambda_{G}\text{Var}[R_{i}](t)
\end{equation}
where $\lambda_{G}$ is a weighting factor which is larger than $1+\lambda_{L}$.

The set of global masks $\mathcal{M}_{i}^{\text{G}}$ of image $I_{i}$ is empty at start:
\begin{equation}
\mathcal{M}_{i}^{G}(0) = \{  \emptyset \}
\end{equation}

Once the residual ${R}_{k}^{j}(t)$ of $m_{k}^{j}$ exceeding the threshold $ T_{k}^{G} $, we mark it as a tracking candidate. Specifically, we select a center point and four edge points of $m_{k}^{j}$  as point prompts, which are then input into Segment Anything Model v2 to initiate tracking. This allows us to obtain the corresponding masks of $m_{k}^{j}$ in each frame:
\begin{equation} 
\mathcal{M}_{i}^{G}(t) =  \mathcal{M}_{i}^{G}(t-1) \cup \hat{m}_{i}^{j}
\end{equation}
where $\hat{m}_{i}^{j}$ denotes the tracking results of $m_{k}^{j}$ in frame $I_{i}$.

The final mask of image $I_{i}$ at iteration $t$ is:
\begin{equation}
\mathcal{M}_{i}(t) = \mathcal{M}_{i}^{L}(t) \cup \mathcal{M}_{i}^{G}(t)
\end{equation}
\subsection{Voxel-guided Gaussian Splatting}
To reconstruct a 3D model with accurate geometry in regions with limited view constraints, we utilize a learning-based multi-view stereo method, DUSt3R \cite{wang2024dust3r}, to obtain rich geometric priors. To integrate these priors within the 3DGS, we design a geometric-aware point sampling method and a voxel-guided Gaussian optimization strategy.

\noindent\textbf{Geometric-aware Point Sampling.}
Given two unposed images, DUSt3R excels at predicting pairwise point maps and confidence maps, as well as recovering camera parameters. 
When processing more than two images, DUSt3R applies a point cloud alignment as post-processing to generate a dense point cloud.
An intuitive idea is to use these dense points to initialize Gaussian primitives. However, the over-parameterized primitives, some of which exhibit positional deviations, make it challenging for 3DGS to achieve effective optimization. To address this issue, we sample dense points based on both confidence and geometric features, ensuring that the sampled points are well-suited for 3DGS initialization while preserving geometric priors.

We employ the Fast Point Feature Histogram(FPFH) Descriptor to extract geometric features from the dense points:
\begin{equation}
\text{FPFH}(p) = \text{SPFH}(p) + \frac{1}{K} \sum_{q \in \mathcal{N}(p)} \frac{1}{d(p, q)} \cdot \text{SPFH}(q)
\end{equation}
where $ \text{SPFH}(p) $ denotes the Simplified Point Feature Histogram for point $ p $, $ \mathcal{N}(p) $ is the set of neighboring points surrounding point $ p $ while $ q $ is a neighbor point within the $ \mathcal{N}(p) $. $ d(p, q) $ is the Euclidean distance between points $ p $ and $ q $ and $ K $ is the number of neighbors considered.

The space occupied by the dense points is then divided into $ N $ voxels, with the voxel size adaptively determined by the scene’s dimensions. In particular,the scene is enclosed within a rectangular cuboid, and the shortest edge of this cuboid is divided into $N$ voxels. For each voxel, we calculate the scores for all points within it:
\begin{equation}
\text{Score}(p) = Conf(p) \cdot \tilde{\text{FPFH}}(p)
\end{equation}
where $Conf(p)$ denotes the confidence of the point $p$ and $\tilde{\text{FPFH}}(p)$ is the normalized FPFH feature of the point $p$.

In each voxel, the top $k$ points with the highest scores are retained to initialize Gaussian primitives.

\noindent\textbf{Voxel-guided Optimization.}
Although these sampled points provide valuable geometric priors for 3DGS, the original optimization strategy could undermine these advantages. Therefore, we additionally incorporate a voxel-guided optimization strategy to direct the optimization of 3DGS under limited view constraints.

For each initialized Gaussian within a voxel, as well as those split or cloned from it, strict constraints are imposed to keep them within the defined bounds of the voxel. 
Specifically, if a Gaussian $ g_i^j $ belonging to voxel $ V_i $ has its center or scale exceed defined limit, typically set to $\tau$ times the voxel length, it will be flagged as an unconstrained Gaussian. The gradient for such Gaussians decays exponentially with increasing distance from the voxel center.
If the accumulated gradient of an unconstrained Gaussian reaches a preset threshold $\gamma_{1}$ and directs toward an empty voxel, it will be split or cloned, and the newly generated Gaussian will be assigned to that empty voxel. 

To eliminate trivial voxels, we remove a voxel if it contains fewer Gaussians than a specified threshold $\gamma_{2}$ or if the average opacity falls below an acceptable level $\gamma_{3}$.
\section{Experiments}

\subsection{Setups}
\label{sec:exp_setups}
\textbf{DroneSplat Dataset.} 
To rigorously evaluate our method in wild scenes, we introduce a novel drone-captured 3D reconstruction dataset. The dataset includes 24 in-the-wild sequences encompassing both dynamic and static scenes, and dynamic scenes containing varying numbers of dynamic distractors. For the experiments of dynamic distractor elimination, we select 6 scenes representing different levels of dynamics, each containing sufficient images to avoid the issue of sparsity; for limited view reconstruction, we select 2 static scenes with only 6 input views. 

\noindent\textbf{On-the-go Dataset.}
The On-the-go dataset \cite{ren2024nerf} includes multiple casually captured scenes with varying ratios of occlusions. Following the baseline settings, we select 6 representative scenes for the distractor elimination experiments.

\noindent\textbf{UrbanScene3D Dataset.}
UrbanScene3D \cite{lin2022capturing} is a large-scale dataset designed for urban scene perception and reconstruction, including both synthetic data and real scenes captured by drones. We select two real scenes with six input views for limited view reconstruction experiments.

\noindent\textbf{Baselines.} For the experiments of distractor elimination, we compare our approach to three NeRF-based methods (RobustNeRF \cite{sabour2023robustnerf}, NeRF-HuGS \cite{chen2024nerf} and NeRF On-the-go \cite{ren2024nerf}) and two 3DGS-based methods (GS-W \cite{zhang2024gaussian} and WildGaussians \cite{kulhanek2024wildgaussians}). In addition, we also include two well-known baselines, the vanilla 3DGS \cite{kerbl3Dgaussians} and Mip-Splatting \cite{yu2024mip}.

For the experiments of limited view reconstruction, we compare two NeRF-based methods (FreeNeRF \cite{yang2023freenerf} and DiffusioNeRF
 \cite{wynn2023diffusionerf}) and three 3DGS-based methods (FSGS \cite{zhu2025fsgs}, DNGaussian \cite{li2024dngaussian} and InstantSplat \cite{fan2024instantsplat}). Moreover, vanilla 3DGS \cite{kerbl3Dgaussians} and a view-adaptive method, Scaffold-GS \cite{lu2024scaffold}, are also included for comparison.

\noindent\textbf{Implementations.}
All baselines are trained following their own default settings. 
For our method, we train the model with 7,000 iterations and all results are obtained using a NVIDIA A100 GPU. 
In the experiments of dynamic distractors elimination, we activate Adaptive Local-Global Masking at 500 iterations, and we set $\lambda_{L}$ to 0.4 while $\lambda_{G}$ to 2.8. 
For geometric-aware point sampling, a resolution of 512 is used for DUSt3R to predict dense points and we set $N$ to 80 with $k$ to 3. 
In Voxel-guided optimization, we set $\tau$ to 3.5, $\gamma_{1}$ to 0.003, $\gamma_{1}$ to 2 and $\gamma_{3}$ to 0.075.

\noindent\textbf{Metrics.} We adopt the widely used PSNR, SSIM \cite{wang2004image} and LPIPS \cite{zhang2018unreasonable} to evaluate the novel view synthesis qualities.

\begin{table*}[t]
  \centering
  \renewcommand{\arraystretch}{1.2} 
  \resizebox{\textwidth}{!}{ 
  \begin{tabular}{
    l|c|c|c|c|c|c|c|c|c|c|c|c|c|c|c|c|c|c
  } 
    \toprule
                     \multirow{3}{*}{Method} & \multicolumn{6}{c|}{\Large Low Dynamic} & \multicolumn{6}{c|}{\Large Medium Dynamic} & \multicolumn{6}{c}{\Large High Dynamic} \\ 
                       & \multicolumn{3}{c|}{\Large Cultural Center} & \multicolumn{3}{c|}{\Large TangTian} & \multicolumn{3}{c|}{\Large Pavilion} & \multicolumn{3}{c|}{\Large Simingshan} & \multicolumn{3}{c|}{\Large Intersection} &  \multicolumn{3}{c}{\Large Sculpture} \\
                     & PSNR$\uparrow$ & SSIM$\uparrow$ & LPIPS$\downarrow$ & PSNR$\uparrow$ & SSIM$\uparrow$ & LPIPS$\downarrow$ & PSNR$\uparrow$ & SSIM$\uparrow$ & LPIPS$\downarrow$ & PSNR$\uparrow$ & SSIM$\uparrow$ & LPIPS$\downarrow$ & PSNR$\uparrow$ & SSIM$\uparrow$ & LPIPS$\downarrow$ & PSNR$\uparrow$ & SSIM$\uparrow$ & LPIPS$\downarrow$ \\ 
    \midrule
    RobustNeRF \small{[CVPR'23]}     &  \large{16.63}   &   \large{0.538}  &  \large{0.541}   &   \large{16.46}  &   \large{0.335}  &  \large{0.589}   &   \large{15.67}  &   \large{0.317}  &  \large{0.496}   &   \large{18.716}  &  \large{0.549}   &   \large{0.441}  &   \large{16.42}  &  \large{0.589}   &  \large{0.444}   &   \large{17.01}  &   \large{0.243}  &  \large{0.602}   \\
    NeRF-HuGS \small{[CVPR'24]}      &  \large{22.02}   &   \large{0.732}  &  \large{0.159}   &   \large{18.01}  &   \large{0.399}  &  \large{0.448}   &   \large{15.90}  &   \large{0.357}  &  \large{0.361}   &   \cellcolor{yellow!30}\large{22.64}  &  \large{0.727}   &   \large{0.187}  &   \large{16.98}  &  \large{0.556}   &  \large{0.407}   &   \large{16.88}  &   \large{0.251}  &  \large{0.459}   \\
    NeRF On-the-go \small{[CVPR'24]}     &  \large{19.53}   &   \large{0.658}  &  \large{0.331}   &   \large{17.59}  &   \large{0.359}  &  \large{0.527}   &   \large{15.58}  &   \large{0.318}  &  \large{0.471}   &   \large{18.128}  &  \large{0.488}   &   \large{0.487}  &   \large{15.76}  &  \large{0.535}   &  \large{0.479}   &   \large{17.04}  &   \large{0.228}  &  \large{0.606}   \\
    \midrule
    3DGS \small{[SIGGRAPH'23]}     &  \large{22.43}   &   \large{0.730}  &  \large{0.152}   &   \large{17.31}  &   \large{0.430}  &  \cellcolor{yellow!30}\large{0.384}   &   \large{17.04}  &   \large{0.438}  &  \large{0.228}   &   \large{22.14}  &  \cellcolor{yellow!30}\large{0.749}   &  \cellcolor{yellow!30}\large{0.127}  &   \large{18.11}  &  \large{0.570}   &  \cellcolor{orange!30}\large{0.317}   &   \large{17.11}  &   \large{0.396}  &  \large{0.256}   \\
    Mip-Splatting \small{[CVPR'24]}     &  \large{22.11}   &  \large{0.727}  &  \large{0.175} &   \large{16.87}  &   \cellcolor{yellow!30}\large{0.436}  &  \large{0.387}   &   \large{17.26}  &   \large{0.441}  &  \cellcolor{red!30}\large{0.210}   &   \large{21.78}  &  \large{0.742}   &   \large{0.137}  &   \large{17.49}  &  \large{0.553}   &  \large{0.346}   &   \cellcolor{yellow!30}\large{17.22}  &   \large{0.404}  &  \cellcolor{yellow!30}\large{0.249}   \\
    GS-W \small{[ECCV'24]}     &  \cellcolor{yellow!30}\large{23.29}   &  \cellcolor{yellow!30}\large{0.759}   &   \cellcolor{yellow!30}\large{0.151}   &   \cellcolor{yellow!30}\large{18.55}  &   \large{0.419}  &  \large{0.528}   &  \large{17.44}   &  \large{0.416}   &  \large{0.424}   &   \large{22.41}  &  \large{0.706}   &  \large{0.243}   &  \cellcolor{yellow!30}\large{19.06}   &   \cellcolor{yellow!30}\large{0.624}  &  \large{0.382}   &  \large{17.09}   &   \cellcolor{yellow!30}\large{0.414}  &  \large{0.270}   \\
    WildGaussians \small{[NIPS'24]}      &   \large{22.41}  &  \large{0.734}   &  \large{0.190}  &  \large{17.45}   &  \large{0.395}   &   \large{0.533}  &  \large{16.96}   &  \large{0.388}   &  \large{0.379}   &   \large{22.63}  &  \large{0.707}   &  \large{0.239}   &  \large{17.22}   &  \large{0.551}  &  \large{0.421}   &  \large{17.15}   &   \large{0.371}  &   \large{0.383}  \\
    Ours(COLMAP)     &  \cellcolor{red!30}\large{24.56}   &   \cellcolor{orange!30}\large{0.773}  &  \cellcolor{red!30}\large{0.142}   &    \cellcolor{orange!30}\large{19.53}  &    \cellcolor{orange!30}\large{0.459}  &  \cellcolor{orange!30} \large{0.365}   &   \cellcolor{red!30}\large{17.89}  &   \cellcolor{orange!30}\large{0.462}  &  \cellcolor{yellow!30}\large{0.217}   &    \cellcolor{orange!30}\large{23.91}  &   \cellcolor{orange!30}\large{0.792}   &   \cellcolor{orange!30}\large{0.098}  &   \cellcolor{orange!30}\large{19.47}  &  \cellcolor{red!30}\large{0.681}   & \cellcolor{red!30} \large{0.304}   &   \cellcolor{orange!30}\large{19.51}  &   \cellcolor{red!30}\large{0.528}  &  \cellcolor{orange!30}\large{0.197}   \\
    Ours       &  \cellcolor{orange!30}\large{24.53}   &   \cellcolor{red!30}\large{0.785}  &  \cellcolor{orange!30}\large{0.149}   &   \cellcolor{red!30}\large{19.68}  &   \cellcolor{red!30}\large{0.472}  &  \cellcolor{red!30}\large{0.331}   &   \cellcolor{orange!30}\large{17.79}  &   \cellcolor{red!30}\large{0.469}  &  \cellcolor{orange!30}\large{0.214}   &   \cellcolor{red!30}\large{24.17}  &  \cellcolor{red!30}\large{0.808}   &   \cellcolor{red!30}\large{0.093}  &   \cellcolor{red!30}\large{19.86}  &  \cellcolor{orange!30}\large{0.657}   &  \cellcolor{yellow!30}\large{0.322}   &   \cellcolor{red!30}\large{19.64}  &   \cellcolor{orange!30}\large{0.517}  &  \cellcolor{red!30}\large{0.192}   \\
    \bottomrule
  \end{tabular}
  }
  \label{table1}
  \vspace{-1em}
\end{table*}

\begin{figure*}[!h]
    \centering
    \includegraphics[width=\textwidth]{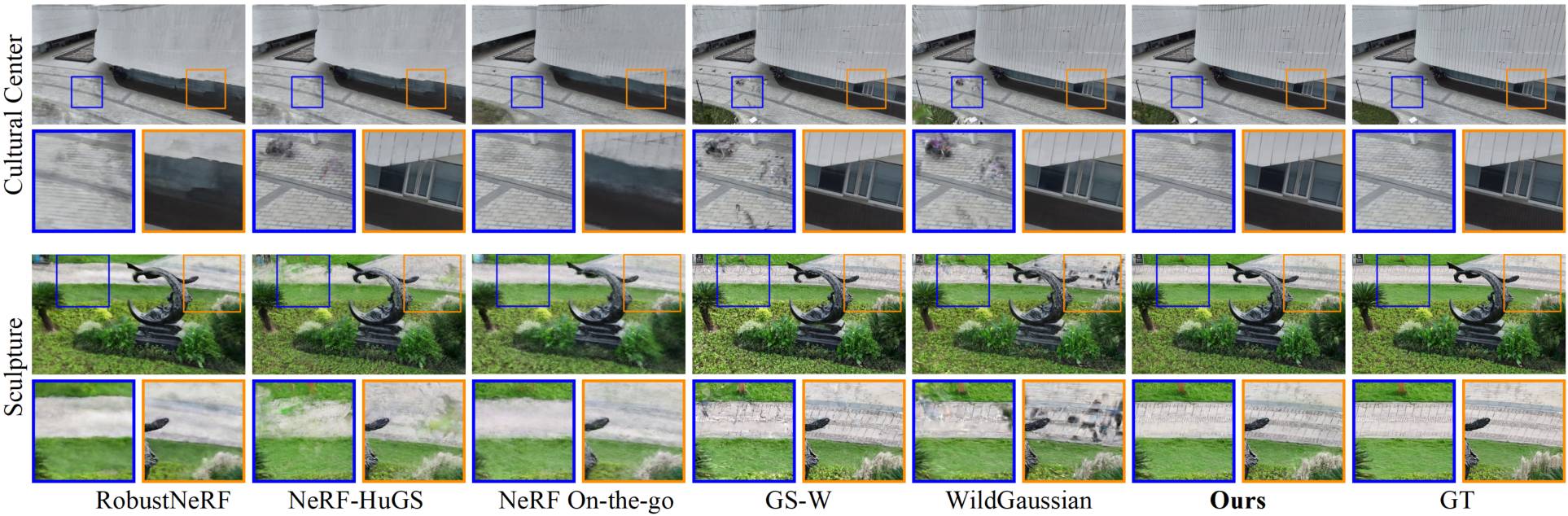}
    \caption{\textbf{Quantitative and qualitative results on DronSplat dataset (dynamic).} The \colorbox{red!30}{1st}, \colorbox{orange!30}{2nd} and \colorbox{yellow!30}{3rd} best results are highlighted. Our method outperforms baseline methods on scenes with various numbers of dynamic distractors, while \textit{Ours(COLMAP)} leading the rest. Note that all scenes have a sufficient number of viewpoints, avoiding the issue of sparsity.}
    \label{compare1}
\end{figure*}

\begin{table*}[!h]
  \centering
  \renewcommand{\arraystretch}{1.2} 
  \resizebox{\textwidth}{!}{ 
  \begin{tabular}{
    l|c|c|c|c|c|c|c|c|c|c|c|c|c|c|c|c|c|c
  } 
    \toprule
                     \multirow{3}{*}{Method} & \multicolumn{6}{c|}{\Large Low Occlusion} & \multicolumn{6}{c|}{\Large Medium Occlusion} & \multicolumn{6}{c}{\Large High Occlusion} \\ 
                     & \multicolumn{3}{c|}{\Large Mountain} & \multicolumn{3}{c|}{\Large Fountain} & \multicolumn{3}{c|}{\Large Corner} & \multicolumn{3}{c|}{\Large Patio} & \multicolumn{3}{c|}{\Large Spot} &  \multicolumn{3}{c}{\Large Patio-High} \\
                     & PSNR$\uparrow$ & SSIM$\uparrow$ & LPIPS$\downarrow$ & PSNR$\uparrow$ & SSIM$\uparrow$ & LPIPS$\downarrow$ & PSNR$\uparrow$ & SSIM$\uparrow$ & LPIPS$\downarrow$ & PSNR$\uparrow$ & SSIM$\uparrow$ & LPIPS$\downarrow$ & PSNR$\uparrow$ & SSIM$\uparrow$ & LPIPS$\downarrow$ & PSNR$\uparrow$ & SSIM$\uparrow$ & LPIPS$\downarrow$ \\ 
    \midrule
    RobustNeRF \small{[CVPR'23]}     &  \large{17.38}   &   \large{0.402}  &  \large{0.523}   &   \large{15.64}  &   \large{0.322}  &  \large{0.561}   &   \large{23.12}  &   \large{0.741}  &  \large{0.186}   &   \large{20.41}  &  \large{0.685}   &   \large{0.197}  &   \large{20.46}  &  \large{0.488}   &  \large{0.458}   &   \large{20.55}  &   \large{0.426}  &  \large{0.407}   \\
    NeRF-HuGS \small{[CVPR'24]}      &  \large{20.21}   &   \cellcolor{yellow!30}\large{0.659}  &  \cellcolor{orange!30}\large{0.160}   &   \large{20.45}  &   \large{0.663}  &  \cellcolor{orange!30}\large{0.167}   &   \large{23.07}  &   \large{0.748}  &  \large{0.191}   &   \large{17.56}  &  \large{0.611}   &   \large{0.261}  &   \large{20.33}  &  \large{0.540}   &  \large{0.361}   &   \large{16.65}  &   \large{0.479}  &  \large{0.397}   \\
    NeRF On-the-go \small{[CVPR'24]}     &  \large{20.07}   &   \large{0.635}  &  \large{0.286}   &   \large{20.23}  &   \large{0.624}  &  \large{0.297}   &   \large{24.02}  &   \large{0.803}  &  \large{0.184}   &   \large{20.65}  &  \large{0.723}   &   \large{0.224}  &   \large{23.31}  &  \large{0.762}   &  \large{0.198}   &   \large{21.44}  &   \large{0.709}  &  \large{0.258}   \\
    \midrule
    3DGS \small{[SIGGRAPH'23]}   &  \large{19.17}   &   \large{0.637}  &  \large{0.176}   &   \large{19.94}  &   \large{0.675}  &  \large{0.172}   &   \large{20.80}  &   \large{0.697}  &  \large{0.211}   &   \large{16.74}  &  \large{0.649}   &   \large{0.201}  &   \large{17.38}  &  \large{0.573}   &  \large{0.426}   &   \large{16.78}  &   \large{0.564}  &  \large{0.346}   \\
    Mip-Splatting \small{[CVPR'24]}   &  \large{18.83}   &   \large{0.616}  &  \large{0.195}   &   \large{19.92}  &   \cellcolor{yellow!30}\large{0.677}  &  \large{0.169}   &   \large{20.26}  &   \large{0.657}  &  \large{0.238}   &   \large{16.19}  &  \large{0.635}   &   \large{0.232}  &   \large{15.59}  &  \large{0.499}   &  \large{0.535}   &   \large{16.47}  &   \large{0.539}  &  \large{0.397}   \\
    GS-W \small{[ECCV'24]}     &  \large{19.92}   &   \large{0.560}  &   \large{0.291}  &  \large{20.19}   &   \large{0.589}  &  \large{0.279}   &  \large{23.72}  &  \large{0.785}   &   \large{0.132}   &  \large{19.10}   &  \large{0.691}   &  \large{0.163}  &  \large{22.42}   &  \large{0.635}   &  \large{0.309}   &  \large{21.21}   &   \large{0.649}  &  \large{0.260}   \\
    WildGaussians \small{[NIPS'24]}   &  \cellcolor{yellow!30}\large{20.43}   &  \large{0.652}   &   \large{0.213}  &  \cellcolor{yellow!30}\large{20.83}   &   \large{0.671}  &   \large{0.185}  &  \cellcolor{yellow!30}\large{24.19}   &   \cellcolor{orange!30}\large{0.816}  &   \cellcolor{yellow!30}\large{0.105}  &  \cellcolor{yellow!30}\large{21.48}    &  \cellcolor{yellow!30}\large{0.806}   &   \cellcolor{yellow!30}\large{0.111}  & \cellcolor{yellow!30}\large{23.76}  &   \cellcolor{yellow!30}\large{0.794}  &  \cellcolor{red!30}\large{0.081}  &  \cellcolor{yellow!30}\large{22.27}  &  \cellcolor{yellow!30}\large{0.732}   &  \cellcolor{red!30}\large{0.165}  \\
    Ours(COLMAP)     &  \cellcolor{orange!30}\large{21.23}   &   \cellcolor{orange!30}\large{0.687}  &  \cellcolor{yellow!30}\large{0.162}   &   \cellcolor{orange!30}\large{21.54}  &  \cellcolor{red!30}\large{0.705}  &  \cellcolor{yellow!30}\large{0.168}   &    \cellcolor{red!30}\large{24.77}  &    \cellcolor{red!30}\large{0.823}  & \cellcolor{yellow!30}\large{0.106}   &   \cellcolor{orange!30}\large{21.85}  &  \cellcolor{red!30}\large{0.816}   &   \cellcolor{orange!30}\large{0.107}  &   \cellcolor{orange!30}\large{24.37}  &  \cellcolor{orange!30}\large{0.821}   &  \cellcolor{yellow!30}\large{0.095}   &   \cellcolor{orange!30}\large{22.53}  &   \cellcolor{orange!30}\large{0.778}  &  \cellcolor{yellow!30}\large{0.181}   \\
    Ours            &  \cellcolor{red!30}\large{21.45}   &   \cellcolor{red!30}\large{0.694}  &  \cellcolor{red!30}\large{0.158}   &   \cellcolor{red!30}\large{21.60}  &   \cellcolor{orange!30}\large{0.699}  &  \cellcolor{red!30}\large{0.165}   &   \cellcolor{orange!30}\large{24.65}  &   \cellcolor{yellow!30}\large{0.814}  &   \cellcolor{red!30}\large{0.098}   &   \cellcolor{red!30}\large{21.88}  & \cellcolor{orange!30} \large{0.812}   &   \cellcolor{red!30}\large{0.106}  &   \cellcolor{red!30}\large{24.44}  &  \cellcolor{red!30}\large{0.827}   &  \cellcolor{orange!30}\large{0.094}   &   \cellcolor{red!30}\large{22.60}  &   \cellcolor{red!30}\large{0.792}  &  \cellcolor{orange!30}\large{0.177}   \\
    \bottomrule
  \end{tabular}
  }
  \label{table2}
  \vspace{-1em}
\end{table*}

\begin{figure*}[!h]
    \centering
    \includegraphics[width=\textwidth]{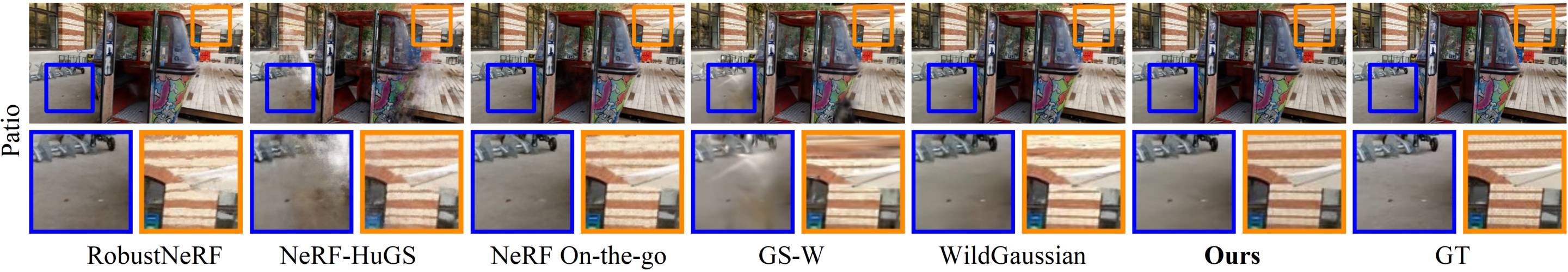}
    \caption{\textbf{Quantitative and qualitative results on NeRF On-the-go dataset.} The \colorbox{red!30}{1st}, \colorbox{orange!30}{2nd} and \colorbox{yellow!30}{3rd} best results are highlighted.  Our method not only effectively eliminates dynamic distractors but also reconstructs fine details in the scene.}
    \label{compare2}
    \vspace{-1em}
\end{figure*}

\subsection{Comparison}
\label{sec:compare}
\noindent\textbf{Distractor Elimination.}
Both the baselines and our method are trained on images with dynamic distractors and evaluated on images without distractors. 

To further validate the effectiveness of Adaptive Local-Global Masking, we introduce a control group, denoted as \textit{Ours (COLMAP)}. This group utilizes sparse point clouds generated by COLMAP for Gaussian initialization, ensuring consistency with other baseline methods, and employs the vanilla 3DGS for subsequent optimization.

As shown in Figure \ref{compare1} and Figure \ref{compare2}, our approach outperforms all baseline method on both DroneSplat(dynamic) datatset and NeRF On-the-go dataset. 
In comparison, GS-W and WildGaussians struggle to eliminate the impact of distractors, leading to artifacts and floaters. While RobustNeRF and NeRF On-the-go successfully remove distractors, they fail to retain fine details.
Our method achieves the highest quantitative results, effectively eliminating dynamic distractors while preserving static details.

\begin{figure}[!t]
    \centering
    \includegraphics[width=\linewidth]{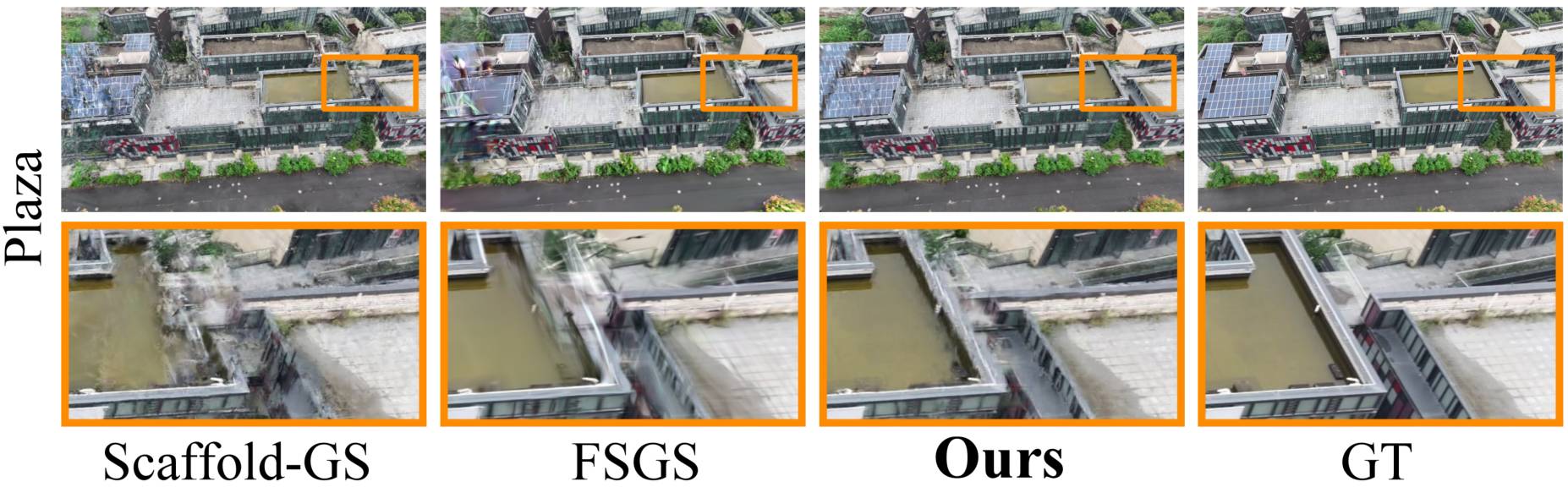}
    \caption{\textbf{Quantitative and qualitative results on DroneSplat dataset(static).} The \colorbox{red!30}{1st}, \colorbox{orange!30}{2nd} and \colorbox{yellow!30}{3rd} best results are highlighted. Our method reconstructs the static scenes with correct geometry even under limited view constraints.}
    \label{compare3}
\end{figure}
\begin{figure}[!t]
    \centering
    \includegraphics[width=\linewidth]{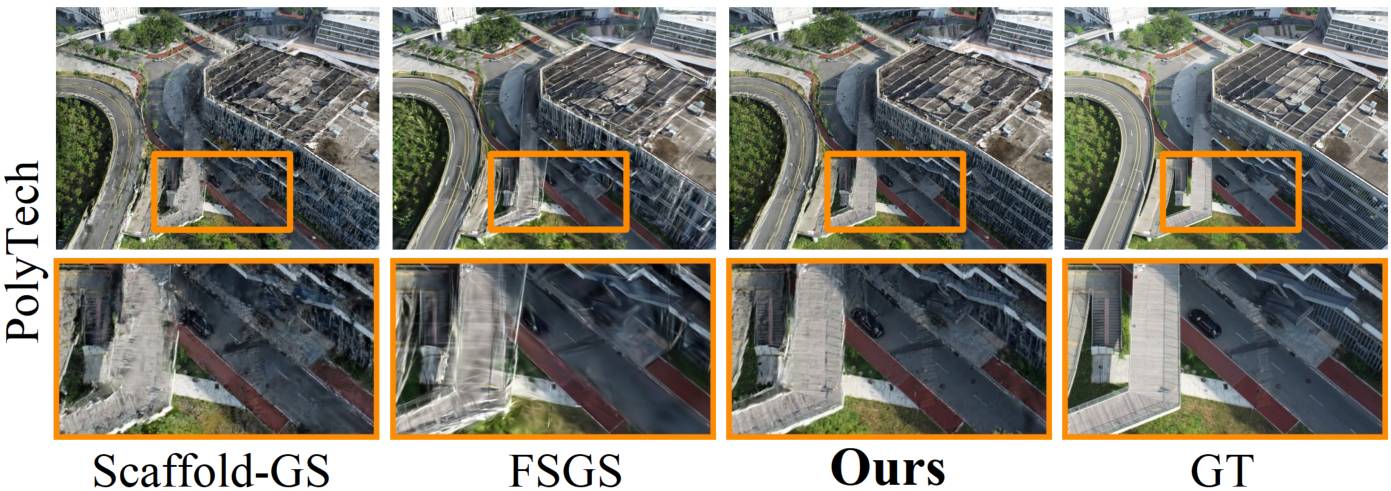}
    \caption{\textbf{Quantitative and qualitative results on UrbanScene3D dataset.} The \colorbox{red!30}{1st}, \colorbox{orange!30}{2nd} and \colorbox{yellow!30}{3rd} best results are highlighted. Our approach shows competitive results compared to state-of-the-art sparse-view reconstruction methods.}
    \label{compare4}
    \vspace{-1em}
\end{figure}

\noindent\textbf{Limited-view Reconstruction.}
As shown in Figure \ref{compare3} and Figure \ref{compare4}, our method demonstrates competitive results on both DroneSplat(static) dataset and UrbanScene3D dataset. 
NeRF-based methods struggle to recover accurate scene geometry. While 3DGS-based methods capture the overall shape, they still generate artifacts due to unconstrained optimization in areas with limited viewpoint overlap (the patches in orange frames).
Despite not specifically designed to address the issue of limited view, Scaffold-GS demonstrates robustness due to its view-adaptive capability. However, when the novel view significantly differs from the input views, the results are still suboptimal.

\begin{table}[!t]
  \centering
  \renewcommand{\arraystretch}{1.05} 
  \resizebox{\linewidth}{!}{ 
  \begin{tabular}{
    l|c|c|c|c|c|c
  } 
    \toprule
                     \multirow{2}{*}{Method}  & \multicolumn{3}{c|}{Exhibition Hall} & \multicolumn{3}{c}{Plaza} \\
                     & PSNR$\uparrow$ & SSIM$\uparrow$ & LPIPS$\downarrow$ & PSNR$\uparrow$ & SSIM$\uparrow$ & LPIPS$\downarrow$   \\ 
    \midrule
    FreeNeRF \small{[CVPR'23]}  &   \large{11.53}   & \large{0.244}   &  \large{0.683}   &   \large{12.30}  &  \large{0.273}    &  \large{0.706} \\
    DiffusioNeRF \small{[CVPR'23]}   &   \large{11.82}   &  \large{0.267}  &  \large{0.656}   &  \large{12.41}   &  \large{0.308}    &  \large{0.695} \\
    \midrule
    3DGS \small{[SIGGRAPH'23]}   &   \large{15.64}   &  \large{0.404}  &  \cellcolor{yellow!30}\large{0.374}   &  \large{16.45}   &   \large{0.438}   &  \cellcolor{yellow!30}\large{0.343} \\
    Scaffold-GS \small{[CVPR'24]}    & \cellcolor{yellow!30}\large{16.55}   &   \cellcolor{orange!30}\large{0.445}  &    \cellcolor{red!30}\large{0.343}  &   \cellcolor{orange!30}\large{17.01}   &   \cellcolor{yellow!30}\large{0.461}   &  \cellcolor{red!30}\large{0.323} \\
    FSGS \small{[ECCV'24]}   &   \large{15.92}   &  \cellcolor{yellow!30}\large{0.424}  &  \large{0.411}   &  \large{16.41}   &  \cellcolor{orange!30}\large{0.469}    &  \large{0.378}  \\
    DNGaussian \small{[CVPR'24]}     &   \large{12.61}   &   \large{0.281}  &  \large{0.558}   &  \large{12.41}   &  \large{0.222}    &  \large{0.595} \\
    InstantSplat  &    \cellcolor{orange!30}\large{17.37}  &  \large{0.393}  &   \large{0.398}  &   \cellcolor{yellow!30}\large{16.61}  &   \large{0.384}   &  \large{0.410} \\
    Ours               &  \cellcolor{red!30}\large{18.09}   &  \cellcolor{red!30}\large{0.496}  &  \cellcolor{orange!30}\large{0.352}   &   \cellcolor{red!30}\large{18.17}  &   \cellcolor{red!30}\large{0.534}   &  \cellcolor{orange!30}\large{0.335} \\
    \bottomrule
  \end{tabular}
  }
  \label{table3}
  \vspace{-1em}
\end{table}

\begin{table}[!t]
  \centering
  \renewcommand{\arraystretch}{1.05} 
  \resizebox{\linewidth}{!}{ 
  \begin{tabular}{
    l|c|c|c|c|c|c
  } 
    \toprule
                     \multirow{2}{*}{Method}  & \multicolumn{3}{c|}{PolyTech} & \multicolumn{3}{c}{ArtSci} \\
                     & PSNR$\uparrow$ & SSIM$\uparrow$ & LPIPS$\downarrow$ & PSNR$\uparrow$ & SSIM$\uparrow$ & LPIPS$\downarrow$   \\ 
    \midrule
    FreeNeRF \small{[CVPR'23]}   &   \large{11.56}   & \large{0.165}   &  \large{0.692}   &   \large{13.42}  &  \large{0.264}    &  \large{0.578} \\
    DiffusioNeRF \small{[CVPR'23]}   &   \large{10.97}   &  \large{0.142}  &  \large{0.769}   &   \large{18.23}  &  \large{0.627}    &  \large{0.235} \\
    \midrule
    3DGS \small{[SIGGRAPH'23]}   &   \large{14.05}   &  \large{0.424}  &  \cellcolor{orange!30}\large{0.279}   &   \large{19.01}  &   \large{0.595}   &  \cellcolor{orange!30}\large{0.201} \\
    Scaffold-GS \small{[CVPR'24]}    &    \cellcolor{yellow!30}\large{14.61}  &   \cellcolor{orange!30}\large{0.479}   &  \cellcolor{red!30}\large{0.253}  &   \cellcolor{yellow!30}\large{19.69}   &   \cellcolor{yellow!30}\large{0.618}   &   \cellcolor{red!30}\large{0.183}  \\
    FSGS \small{[ECCV'24]}     &   \cellcolor{orange!30}\large{14.82}   &   \cellcolor{yellow!30}\large{0.473}  &   \large{0.307}   &  \cellcolor{orange!30}\large{20.32}  &  \cellcolor{red!30}\large{0.651}    &  \large{0.240} \\
    DNGaussian \small{[CVPR'24]}    &    \large{12.99}  &    \large{0.347}  &  \large{0.297}  &  \large{15.43}   &  \large{0.439}   &  \large{0.382}   \\
    InstantSplat      &   \large{14.51}   &  \large{0.365}  &   \large{0.337}  &   \large{16.16}  &  \large{0.325}    &  \large{0.364} \\
    Ours               &   \cellcolor{red!30}\large{15.62}   &  \cellcolor{red!30}\large{0.516}  &   \cellcolor{yellow!30}\large{0.288}  &  \cellcolor{red!30}\large{20.47}   &   \cellcolor{orange!30}\large{0.639}   &  \cellcolor{yellow!30}\large{0.215} \\
    \bottomrule
  \end{tabular}
  }
  \label{table4}
  \vspace{-1em}
\end{table}

\subsection{Ablation Study}
\begin{figure}[!t]
    \centering
    \includegraphics[width=\linewidth]{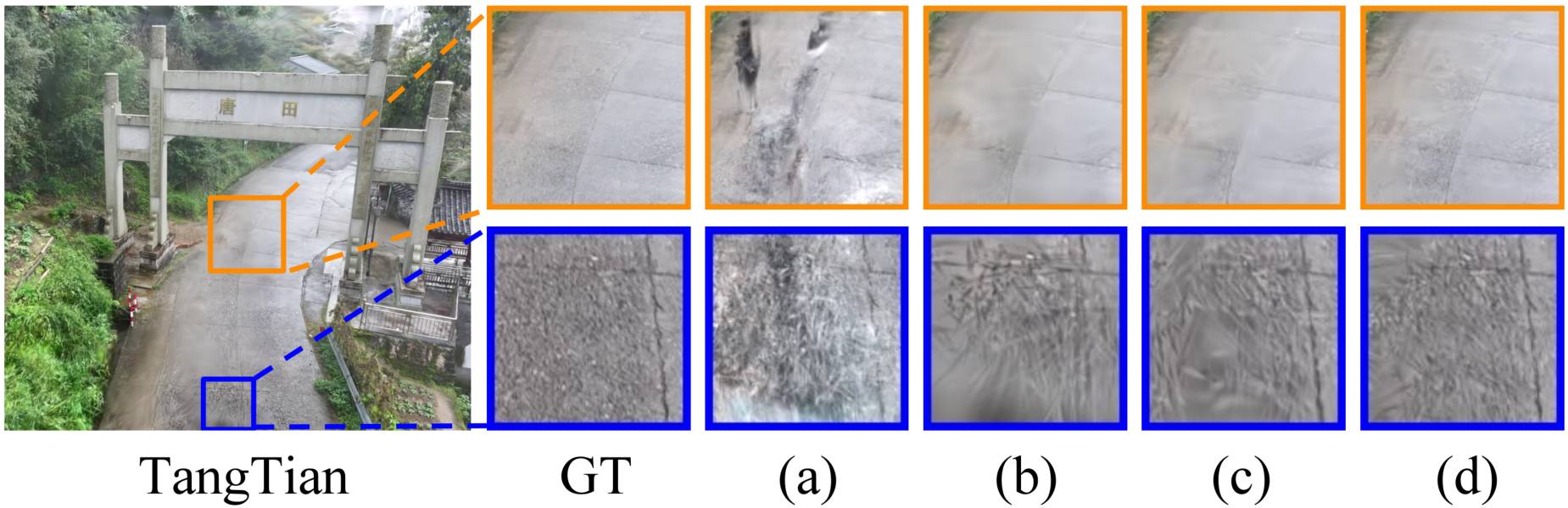}
    \caption{\textbf{The ablations of our method on the DroneSplat(dynamic) and NeRF On-the-go dataset.} The \colorbox{red!30}{1st}, \colorbox{orange!30}{2nd} and \colorbox{yellow!30}{3rd} best results are highlighted.}
    \label{ablation1}
\end{figure}

We divide our method into three modules to analyze their contributions. As shown in Figure \ref{ablation1}, 
Adaptive Local Masking effectively identifies and eliminates distractors in dynamic scenes (b) while Complementary Global Masking serves as an additional check for any missed distractors (c).
Incorporating Voxel-guided 3DGS further enhances performance (d). This improvement is attributed to our voxel-guided optimization, which constrains the spread of distractors and removes voxels with a high proportion of distractors by calculating the average opacity within each voxel.

\begin{figure}[!t]
    \centering
    \includegraphics[width=\linewidth]{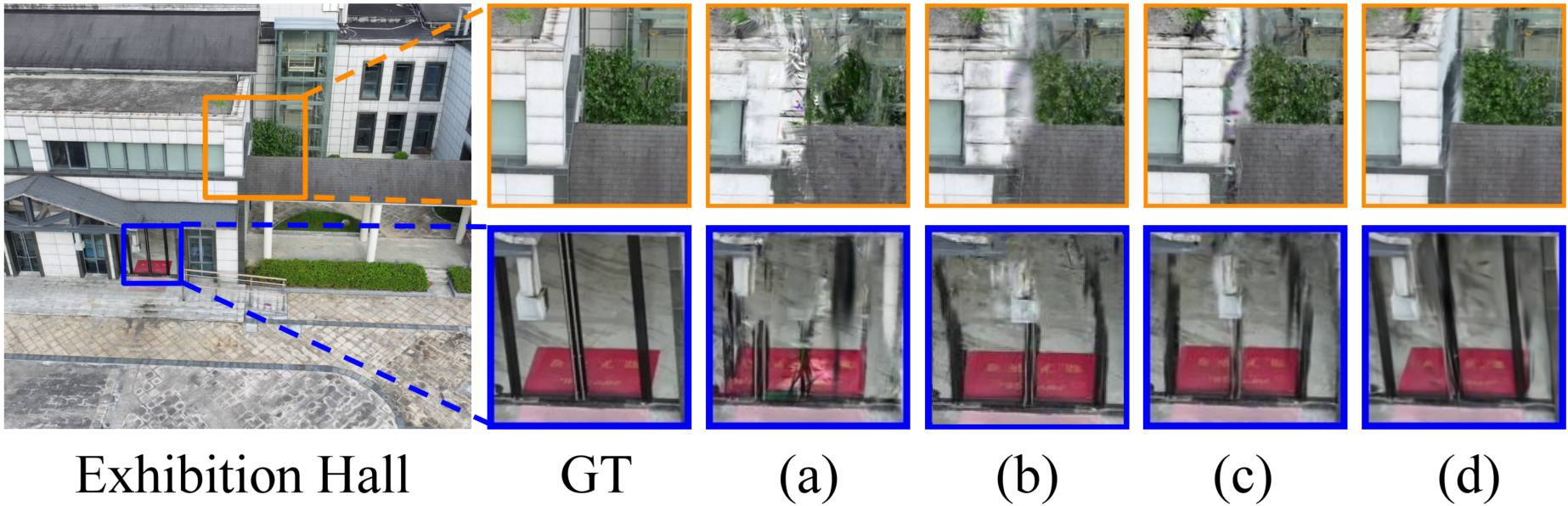}
    \caption{\textbf{The ablations of Voxel-guided 3DGS on the Dronesplat(static) and UrbanScene3D dataset. } The \colorbox{red!30}{1st}, \colorbox{orange!30}{2nd} and \colorbox{yellow!30}{3rd} best results are highlighted.}
    \label{ablation2}
    \vspace{-1em}
\end{figure}

To further study the Voxel-guided 3DGS, we remove different components to verify their effect. As shown in Figure \ref{ablation2}, each component plays a critical role. Notably, despite leveraging geometric priors from multi-view stereo, the reconstruction quality does not improve as expected when using vanilla 3DGS, indicating the vanilla 3DGS struggles to optimize the over-parameterization initial primitives (b). The sampled points retain the prior of scene geometry, yet subsequent Gaussian optimization undermines the prior (c). The complete method that combines the prior and voxel-guided  optimization strategy achieves the best results (d).

\begin{table}[!t]
  \centering
  \renewcommand{\arraystretch}{1.2} 
  \resizebox{\linewidth}{!}{ 
  \begin{tabular}{
   lccc|ccc|ccc
  } 
    \toprule
                     & Local  &  Global &  Voxel-guided &  \multicolumn{3}{c|}{DroneSplat(dynamic)}  &  \multicolumn{3}{c}{NeRF On-the-go} \\ 
                     & Masking &   Masking &  3DGS & PSNR$\uparrow$  & SSIM$\uparrow$ &  LPIPS$\downarrow$  & PSNR$\uparrow$  & SSIM$\uparrow$ &  LPIPS$\downarrow$\\
    \midrule
      (a) & \ding{55} &   \ding{55}  & \ding{55}   &  \large{19.02}   &   \large{0.552}  &  \large{0.244}  &  \large{18.46}  &  \large{0.632}   &  \large{0.255}  \\
      (b) & \checkmark &  \ding{55}  & \ding{55}   &  \cellcolor{yellow!30}\large{20.71}   &   \cellcolor{yellow!30}\large{0.609}  &  \cellcolor{yellow!30}\large{0.223}   &   \cellcolor{yellow!30}\large{22.64}  &  \cellcolor{yellow!30}\large{0.764}   &  \cellcolor{yellow!30}\large{0.140}\\
      (c) & \checkmark &  \checkmark   &   \ding{55}  &  \cellcolor{orange!30}\large{20.74}   &  \cellcolor{orange!30}\large{0.616}   &  \cellcolor{orange!30}\large{0.221}   &   \cellcolor{orange!30}\large{22.71}   &  \cellcolor{orange!30}\large{0.772}   &  \cellcolor{orange!30}\large{0.136} \\
      (d) & \checkmark &  \checkmark   &   \checkmark  &  \cellcolor{red!30}\large{20.86}  &  \cellcolor{red!30}\large{0.618}   &  \cellcolor{red!30}\large{0.217}   &   \cellcolor{red!30}\large{22.77}  &   \cellcolor{red!30}\large{0.773}   &  \cellcolor{red!30}\large{0.133} \\
    \bottomrule
  \end{tabular}
  }
  \vspace{-1em}
\end{table}

\begin{table}[!t]
  \centering
  \renewcommand{\arraystretch}{1.2} 
  \resizebox{\linewidth}{!}{ 
  \begin{tabular}{
   lccc|ccc|ccc
  } 
    \toprule
                     & Geometric &   Point  &  Gaussian &     \multicolumn{3}{c|}{DroneSplat(static)}  &  \multicolumn{3}{c}{UrbanScene3D} \\ 
                     & Priors  & Sampling &  Optimization &    PSNR$\uparrow$  & SSIM$\uparrow$ &  LPIPS$\downarrow$  & PSNR$\uparrow$  & SSIM$\uparrow$ &  LPIPS$\downarrow$\\
    \midrule
     (a) & \ding{55} & \ding{55} &   \ding{55}  &  \large{16.01}  &  \cellcolor{yellow!30}\large{0.421}  &   \cellcolor{orange!30}\large{0.358}  &  \cellcolor{yellow!30}\large{16.53}  &  \cellcolor{yellow!30}\large{0.510}   &  \cellcolor{red!30}\large{0.241}  \\
     (b) &  \checkmark & \ding{55} &  \ding{55}  &   \cellcolor{yellow!30}\large{16.99}  &   \large{0.388}  &   \large{0.404}  &  \large{15.33}  &  \large{0.345}   &  \large{0.351}  \\
     (c) &  \checkmark & \checkmark & \ding{55}  &  \cellcolor{orange!30}\large{17.46}    &   \cellcolor{orange!30}\large{0.432}  &  \cellcolor{yellow!30}\large{0.359}   &  \cellcolor{orange!30}\large{16.97}  & \cellcolor{orange!30}\large{0.536}   &   \cellcolor{yellow!30}\large{0.289} \\
    (d) & \checkmark & \checkmark &  \checkmark   &  \cellcolor{red!30}\large{18.13}  &   \cellcolor{red!30}\large{0.515}  &   \cellcolor{red!30}\large{0.343}  &  \cellcolor{red!30}\large{18.05}  &   \cellcolor{red!30}\large{0.578}  &  \cellcolor{orange!30}\large{0.252}  \\
    \bottomrule
  \end{tabular}
  }
  \vspace{-1em}
\end{table}

\section{Conclusions}
We present DroneSplat, a  novel framework for robust 3D reconstruction from in-the-wild drone imagery.
We integrate local-global segmentation  heuristics and statistical approaches to precisely identify and eliminate dynamic distractors.
Furthermore, We enhance 3D Gaussian Splatting through multi-view stereo predictions and a voxel-guided optimization strategy, enabling accurate scene reconstruction under limited view constraints.
Experimental evaluations across diverse datasets demonstrate the superiority and robustness of our approach over previous methods.
\section*{Acknowledgements}
This work was partly supported by National Natural Science Foundation of China (Grant No. NSFC 62233002) and National Key R\&D Program of China (2022YFC2603600). 
The authors would like to thank Tianji Jiang, Xihan Wang, and all other members of ININ Lab of Beijing Institute of Technology for their contribution to this work.
{
    \small
    \bibliographystyle{ieeenat_fullname}
    \bibliography{main}
}

\clearpage
\setcounter{page}{1}
\maketitlesupplementary

\section{Implementation Details}
\label{sec:details}
\subsection{Datasets}
\textbf{DroneSplat Dataset.} DroneSplat Dataset is acquired with a DJI Mavic Pro 3 drone. The drone-captured images have two resolutions: 1920 $\times$ 1080 and 3840 $\times$ 2160. The dataset contains 24 in-the-wild drone-captured sequences, encompassing both dynamic and static scenes.
The dynamic scenes feature a diverse range of moving objects, including cars, trucks, tricycles, pedestrians, strollers, wind-blown flags, etc. 
Furthermore, dynamic scenes are categorized into three levels based on the number of dynamic objects in the training set: "low dynamic" indicates scenes with only 4-10 dynamic objects, "medium dynamic" includes 10-50 dynamic objects, and "high dynamic" refers to scenes containing more than 50 dynamic objects.

\begin{figure*}[!h]
    \centering
    \includegraphics[width=\textwidth]{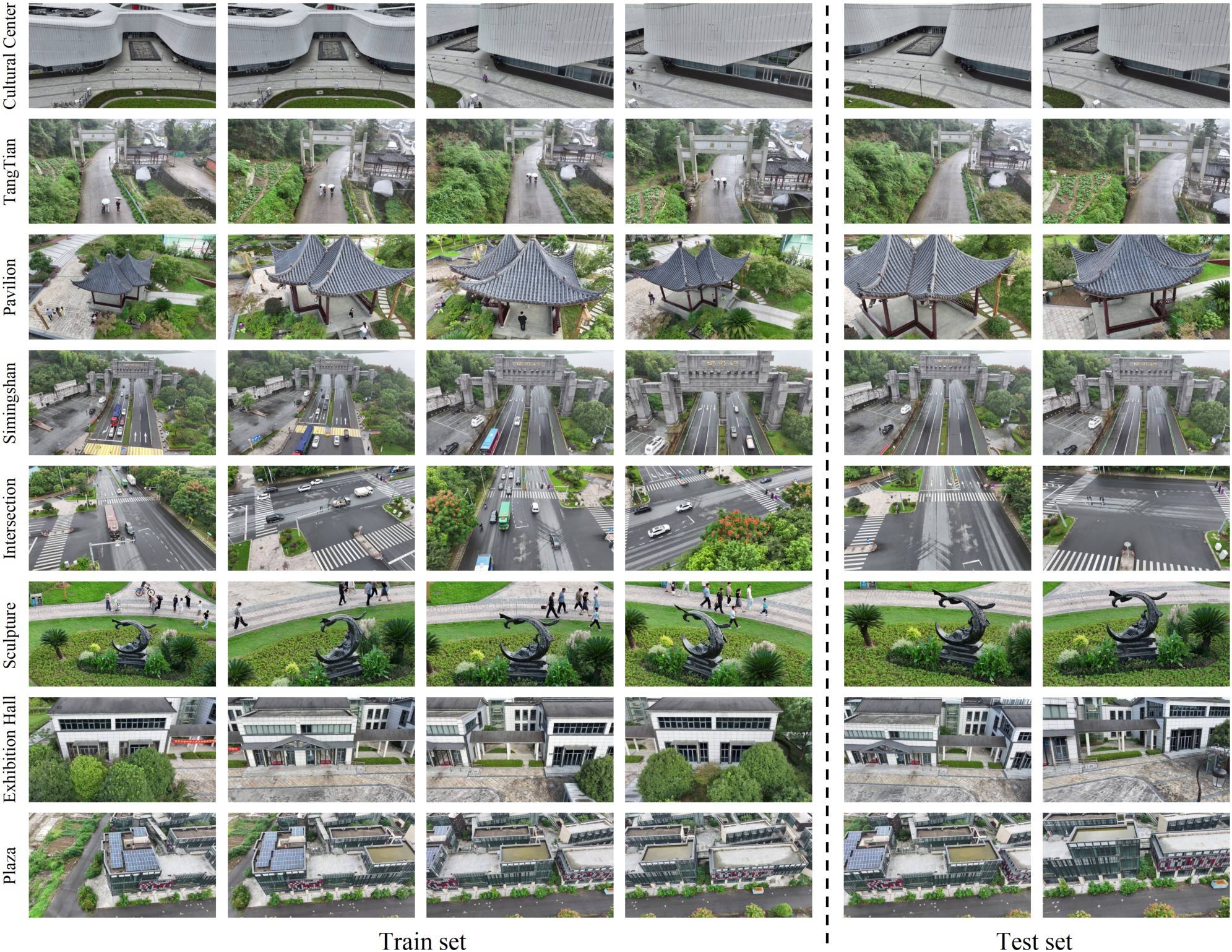}
    \caption{\textbf{DroneSplat Dataset.} Our dataset is captured in an uncontrolled wild environment by a drone, distinguishes itself from prior 3D reconstruction datasets by featuring numerous dynamic distractors with small area in each scene.}
    \label{drone_dataset}
\end{figure*}

Although previous 3D reconstruction datasets also have drone-captured scenes, such as the \textit{drone} in NeRF On-the-go, their test images still contain dynamic distractors, which can introduce ambiguity in metric evaluation. In contrast, the test images in DroneSplat dataset's dynamic scenes feature only static elements, enabling a more rigorous and precise evaluation of our method and the baselines (Figure \ref{drone_dataset}).

The dataset will be released soon on our project page.

\noindent\textbf{NeRF On-the-go Dataset \cite{ren2024nerf}} The NeRF On-the-go dataset comprises 12 casually captured sequences, featuring 10 outdoor scenes and 2 indoor scenes, with occlusion ratios ranging from 5\% to over 30\%.
The images in this dataset are available in two resolutions: most are 4032 $\times$ 3024 and a few are 1920 $\times$ 1080. To accelerate model training, NeRF On-the-go downsamples these resolutions by a factor of eight and four, respectively, in its experiments.

In the NeRF On-the-go dataset, dynamic distractors are typically limited in number (usually the data collector's companions) but may occupy a significant portion of the image. In contrast, most scenes in the DroneSplat dataset contain numerous dynamic distractors (like cars on the road), but the proportion of dynamic distractors in the image is relatively small. 
To validate the effectiveness and robustness of our Adaptive Local-Global Masking method on real-world data with diverse characteristics, we perform dynamic distractor elimination on six scenes from this dataset.

\subsection{Adaptive Local-Global Masking.}
\noindent\textbf{Adaptive Local Masking.} 
The role of Adaptive Local Masking is to identify dynamic distractors in the training image $I_{t}$ at the current iteration $t$. Since the residuals across different scenes and iterations can vary significantly, we adaptively adjust the masking threshold based on the current Object-wise Average of Normalized Residuals and statistical approaches.

\begin{figure}[!t]
    \centering
    \includegraphics[width=\linewidth]{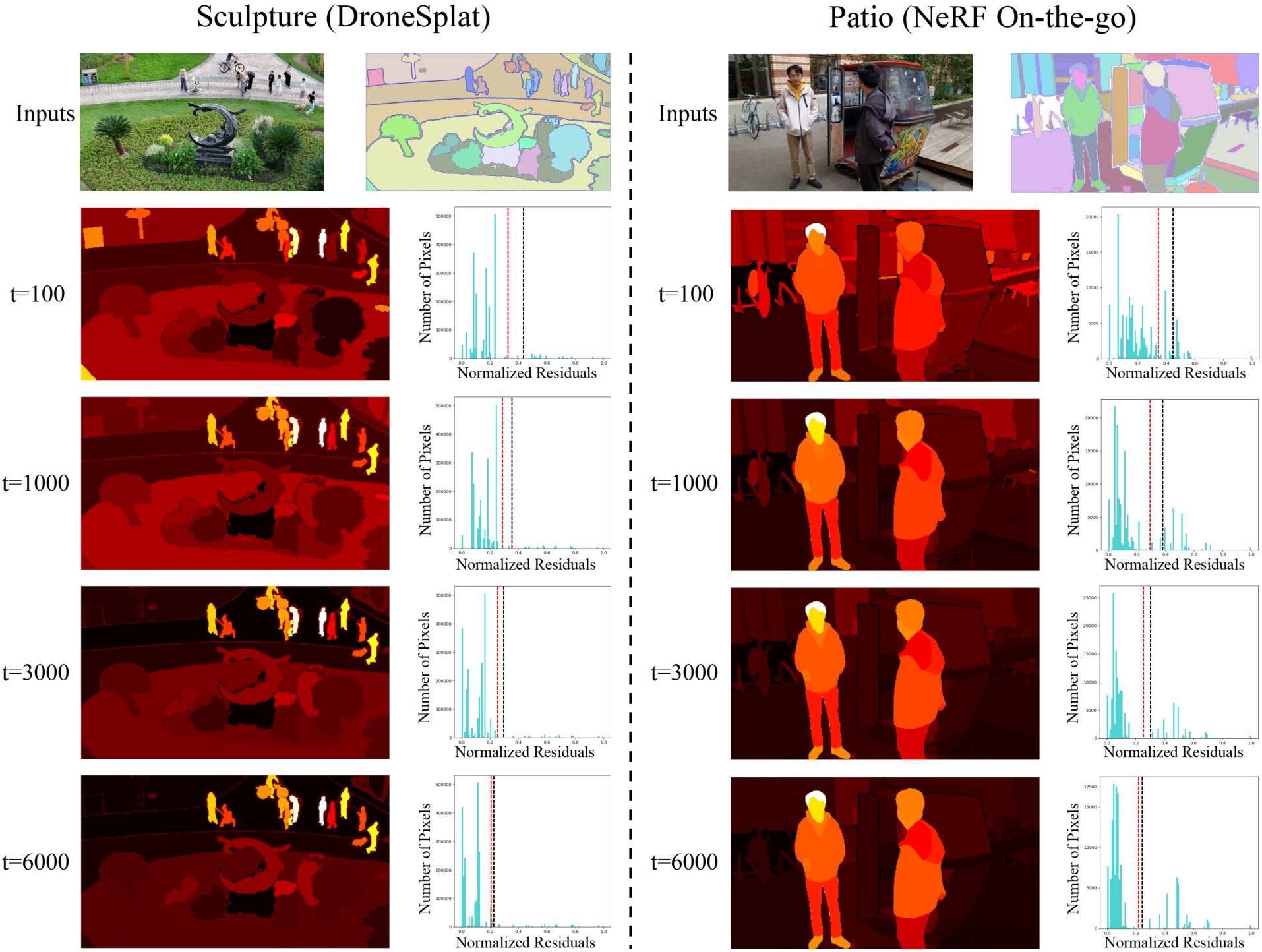}
    \caption{\textbf{Adaptive mask thresholding.} The first row of images shows the input training image and its corresponding segmentation result. The subsequent rows illustrate the Object-wise Average of Normalized Residuals and their corresponding histograms at different iterations $t$. The red dashed line in the histograms represents the sum of the current normalized residual's mathematical expectation and one standard deviation. The black dashed line represents the local masking threshold finally selected.}
    \label{supp_localmask}
    \vspace{-1em}
\end{figure}

A key observation is that, for the same image, the residuals of the static scene gradually decrease as $ t $ increases while the residuals of dynamic distractors remain almost unchanged. Furthermore, statistical analysis shows that both the mathematical expectation and variance of the Object-wise Average of Normalized Residuals for the image also decrease over time.  
If dynamic distractors do not dominate the scene, the mathematical expectation of the Object-wise Average of Normalized Residuals declines rapidly as the static scene converges. However, the presence of dynamic distractors causes the variance to decrease at a much slower rate than the expectation.

As shown in Figure \ref{supp_localmask}, whether in the DroneSplat dataset, characterized by numerous small-area dynamic objects, or the NeRF On-the-go dataset, with fewer but larger dynamic objects, nearly all static objects remain within one standard deviation of the expectation, regardless of whether it is the early training stage with higher residuals or the later stage as residuals converge.

\noindent\textbf{Complement Global Masking.} 
The role of Complement Global Masking is to identify the corresponding masks in the context of the object with particularly high residual in the current training frame. Specifically, we mark the objects with residuals higher than the threshold $\mathcal{T}^{G}$ in the Object-wise Average of Normalized Residuals as tracking candidates (there can be multiple candidates in the same iteration).
For each candidate, several points are selected as prompts and fed into Segment Anything Model v2 \cite{ravi2024sam} for video segmentation.

\begin{figure*}[!t]
    \centering
    \includegraphics[width=\textwidth]{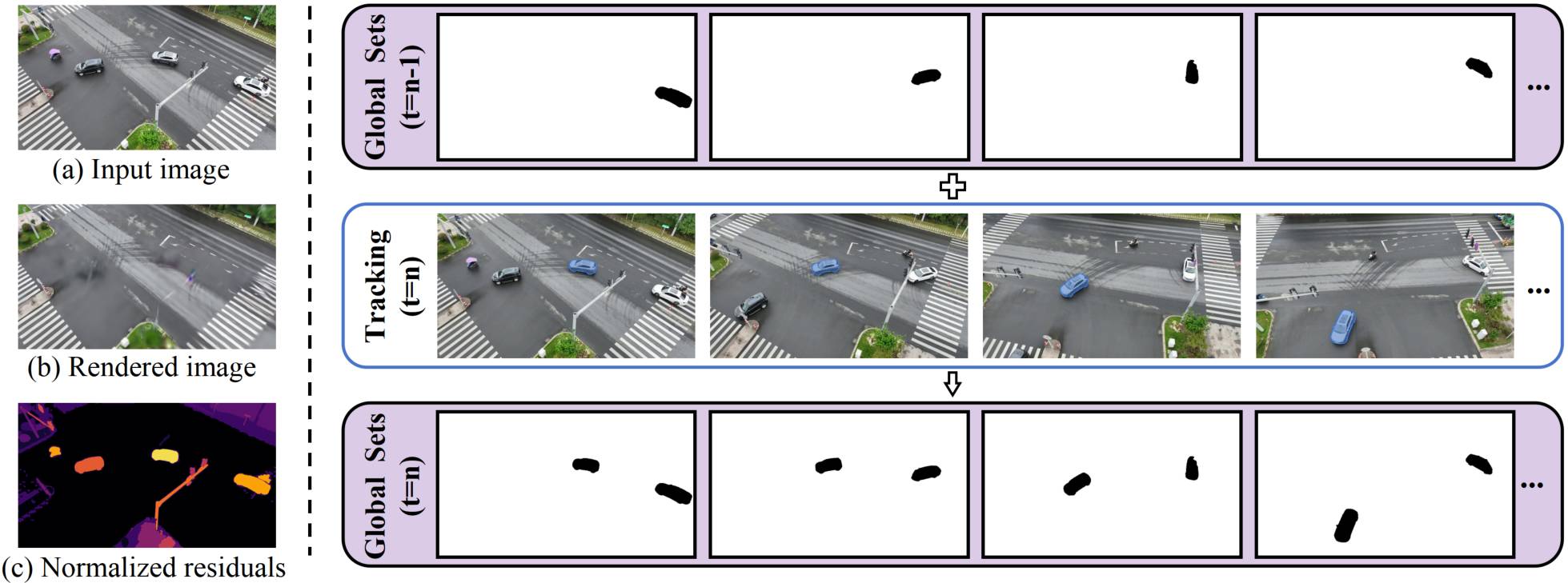}
    \caption{\textbf{Complement Global Masking.} At $t = n$, the white vehicle in the center of the image is marked as a tracking candidate due to its residual exceeding the global masking threshold. The tracking results (blue masks) are then incorporated into the global sets from the previous iteration $t=n-1$ to update the global sets for the current iteration.}
    \label{supp_globalmask}
\end{figure*}

As shown in Figure \ref{supp_globalmask}, we can obtain the high residual objects (the white car which is highlighted in yellow in the residual image) that needs to be tracked based on the Object-wise Average of Normalized Residuals and the predefined global masking threshold.
The blue box on the right represents the tracking results, specifically the mask of the tracked vehicle within the context. 
By combining these tracking results with the global sets from the previous iteration, we can obtain the updated global sets for the current iteration.

Complement Global Masking is designed to address specific cases that Adaptive Local Masking cannot handle. For example, as discussed in the main paper, when a vehicle stops at a red light at an intersection, Adaptive Local Masking may fail to identify the stationary vehicle as a dynamic  distractor in those frames. Unsurprisingly, we observe that lowering the global masking threshold $\mathcal{T}^{G}$ enables Complement Global Masking to independently and effectively eliminate dynamic distractors. However, due to the significant time cost of video segmentation, relying solely on Complement Global Masking for dynamic object identification would prolong the training process. Therefore, the optimal performance is achieved by combining Adaptive Local Masking with Complement Global Masking.

\subsection{Voxel-guided Gaussian Splatting.}
\noindent\textbf{Multi-view Stereo.} 
DUSt3R \cite{wang2024dust3r} is a learning-based framework that takes image pairs as input and outputs corresponding dense point clouds. A post-processing is then used to align the scale across different pairs and obtain a global point cloud. However, as the number of input images increases, the number of image pairs also grows, significantly increasing GPU memory consumption during post-processing. Consequently, the vanilla DUSt3R framework is not well-suited for handling in-the-wild scenes with a large number of images.

To address this issue, we optimize the pipeline by introducing a progressive alignment strategy. Specifically, we divide all images into batches. Suppose there are $b \times N$ images in total, divided into $b$ batches, each containing $N$ images. We first input the $N$ images from the first batch into DUSt3R to generate pair-wise point clouds, followed by post-processing to obtain the aligned point clouds (denoted as $P_{1}$) and the corresponding camera parameters $C_{1}$. Next, we take the last $N/2$ images from the first batch and the first $N/2$ images from the second batch, inputting them into DUSt3R for pair-wise point cloud prediction, where the poses of the first $N/2$ images are fixed using $C_{1}$. Through post-processing, we obtain the aligned point clouds for the second batch (denoted as $P_{2}$) and the camera parameters $C_{2}$. Note that $P_{1}$ and $P_{2}$, as well as $C_{1}$ and $C_{2}$, share $N/2$ overlapping elements, so the duplicated $N/2$ must be removed when merging the outputs of the two batches. Each batch undergoes the aforementioned process, finally resulting in a globally aligned point cloud assembled from multiple batches, along with all corresponding camera parameters. Incorporating the progressive alignment strategy, DUSt3R can handle any number of images by adjusting the number of images per batch.

\noindent\textbf{Geometric-aware Point Sampling.}
The multi-view stereo method provides rich scene geometry priors, and the number of points in the dense point cloud produced by DUSt3R is still very large even with limited viewpoints. 
Taking a resolution of 1920 $\times$ 1080 as an example, the images will be automatically resized to 512 $\times$ 288 when input into DUSt3R, and using just six images yields a point cloud containing over 800,000 points. 
Directly using such a large number of points to initialize Gaussian primitives makes the vanilla optimization strategy ineffective, as each pixel influenced by an excessive number of similar Gaussians, leading to suboptimal reconstruction quality.

Therefore, we propose a geometric-aware point cloud sampling method. 
The entire scene is divided into smaller voxels, and for each voxel, only a certain number of points are retained for 3DGS initialization, selected based on their geometric features and confidence score.
As shown in Figure \ref{supp_point}, the sampled point cloud retains the geometric priors, preserving the scene's overall structure. Meanwhile, the number of points is significantly reduced, from over 800,000 to fewer than 100,000. This reduction allows the 3DGS's densification process to fully leverage its remarkable representational capabilities. Notably, the importance of the point sampling method becomes particularly evident when the number of input images increases and overlap regions expand. In such cases, DUSt3R's output may exhibit aliasing, for example, there are many layers of close-fitting ground and walls,, which can severely impact 3DGS's performance. The sampling method effectively mitigates these issues, ensuring high-quality initialization.

\begin{figure}[!h]
    \centering
    \includegraphics[width=\linewidth]{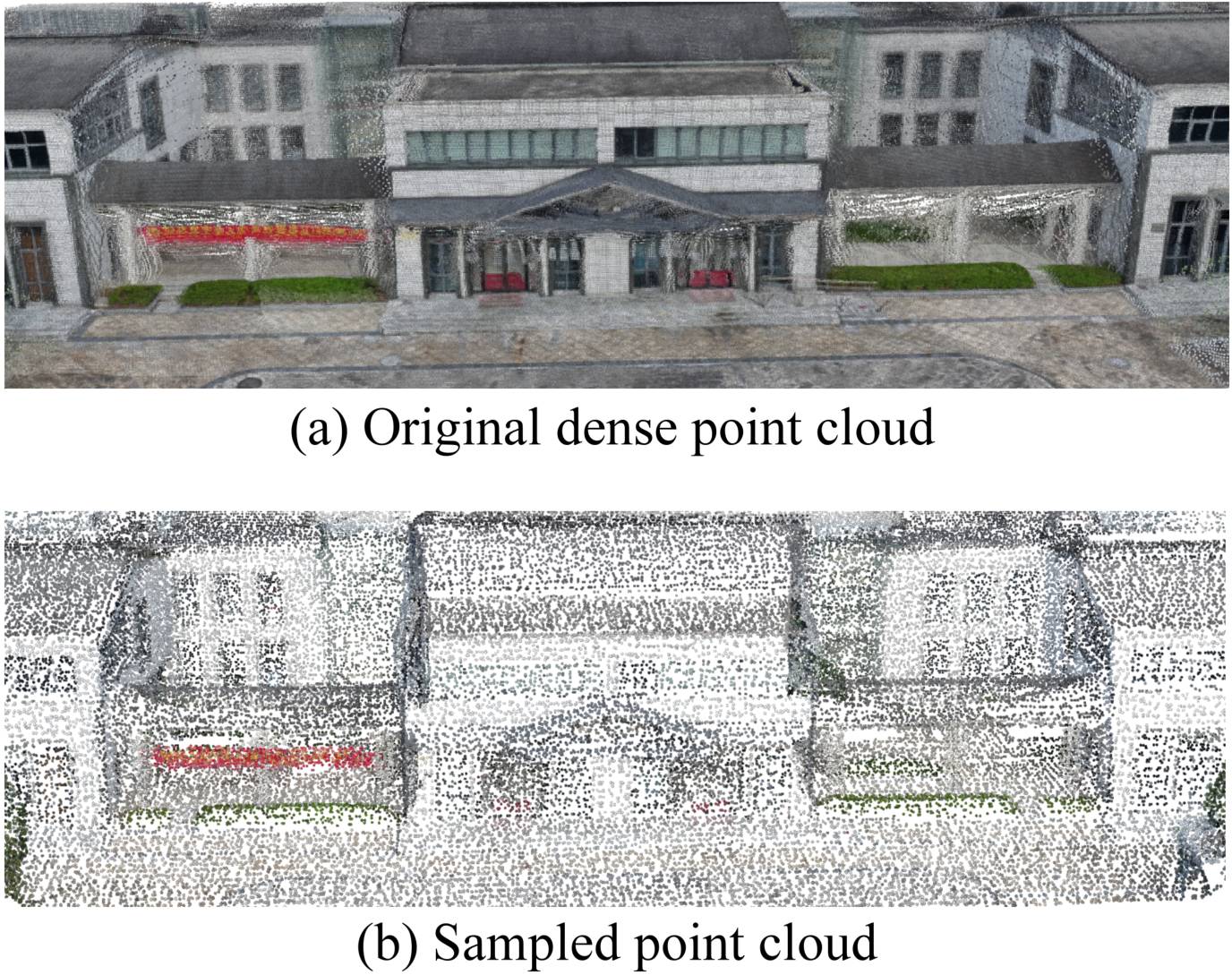}
    \caption{\textbf{The effect of point sampling.}  Compared to the original point cloud (a), the sampled point cloud (b) also provides sufficient geometric priors using only a fraction of the points.}
    \label{supp_point}
    \vspace{-1em}
\end{figure}

\noindent\textbf{Voxel-guided Optimization.}
The role of Voxel-guided Optimization is to overcome the challenge of unconstrained optimization in 3DGS under limited viewpoints. As illustrated in Figure \ref{supp_point}, when a drone flies over a street with a fixed posture, the angle between the camera and the ground remains nearly constant (a). Using the original 3DGS optimization results in uncontrolled Gaussian expansion and drifting. Specifically, the unconstrained Gaussian moves toward the camera, which will cause floating Gaussians in the air (b). However, this does not affect optimization for the training views, it highlights a limitation: the Gaussians lack sufficient constraints under limited viewpoints when relying solely on the original optimization.

Our proposed Voxel-guided Optimization addresses the issue of unconstrained optimization by restricting Gaussians within a defined space. Leveraging the rich geometric priors provided by multi-view stereo, we achieve basic overlay of the scene in Gaussian initialization. As mentioned in the main paper, Gaussians that exceed the voxel restricted boundary are identified as unconstrained and have their gradients reduced. Figure \ref{supp_point} (c) demonstrates the effectiveness of our approach.
There are no more floating Gaussians in the air, and the surface of the building is reconstructed more accurately. Notably, Voxel-guided Optimization does not interfere with the original optimization process.

Among all the hyperparameters in Voxel-guided Optimization, the most critical is $\tau$, which controls the voxel constraint boundary. A value of $\tau$ that is too small prevents Gaussians from adequately fitting the scene, while a value that is too large weakens the voxel's constraint on Gaussian optimization. In practice, we find that setting $\tau$ between 3 and 4 strikes an effective balance, enabling accurate Gaussian fitting while maintaining sufficient constraints.

\begin{figure}[!h]
    \centering
    \includegraphics[width=\linewidth]{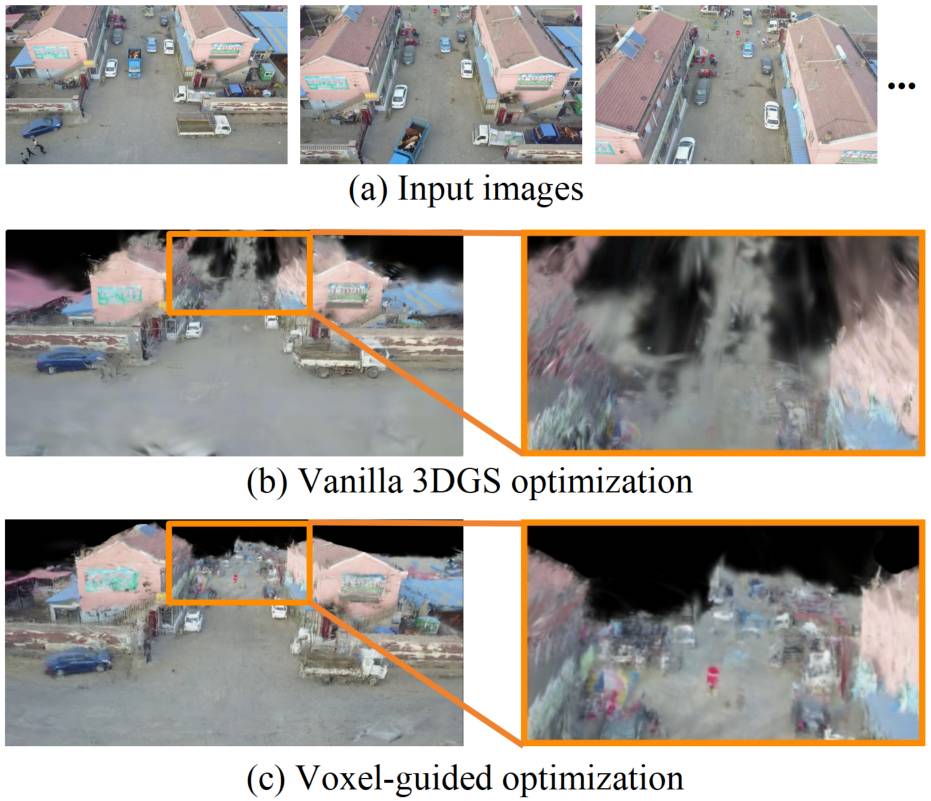}
    \caption{\textbf{The effect of Voxel-guided Optimization.} Compared to the vanilla optimization, our voxel-based optimization strategy ensure accurate scene geometry. Notably, both (b) and (c) represent the visualization of Gaussians, not the rendering results.}
    \label{supp_point}
    \vspace{-1em}
\end{figure}

\section{Additional Experiments}
\subsection{Ablation Study}
As shown in Table \ref{dense_point}, dense point clouds provide a strong geometric prior, accelerating convergence and reducing overall reconstruction time. While initialization of dense points requires more memory, final memory usage and model size remain comparable to sparse point clouds.

As shown in Table \ref{gpu_time}, among all the designed modules, local masking requires the longest processing time while also providing the most significant performance improvement. Furthermore, although the training time on the 3090 is slightly longer compared to the A100 and 4090, it remains within an acceptable range.

\begin{table}[!h]
\centering
\setlength{\tabcolsep}{5pt}
\caption{\textbf{Impact of initial points on memory Usage, training time, model size, and PSNR.} We conduct the experiments on DroneSplat (static) dataset. For each scene, row 1: COLMAP; row 4: DUSt3R; rows 2-3: DUSt3R downsampled points.}
\label{dense_point}
\small
\begin{tabular}{cccccc}
\toprule
\multirow{2}{*}{Scene} & Init & Memory & Training & Final & \multirow{2}{*}{PSNR$\uparrow$} \\
& Points & Usage & Time & Size & \\
\midrule
\multirow{4}{*}{Hall} & 15426 & 8.17GB & 7.72m & 667.69MB & 15.61 \\
 & 47489 & 8.32GB & 6.73m & 729.29MB &  16.48 \\
 & 237446 & 8.67GB & 6.53m & 802.16MB  & 17.92\\
 & 1187233 & 9.21GB & 7.33m & 973.62MB & 17.63  \\
\midrule
\multirow{4}{*}{Plaza} & 11991 & 6.74GB & 6.67m & 481.16MB & 16.38\\
 & 43885 & 6.03GB & 5.75m & 496.41MB & 16.74 \\
 & 219429 & 6.12GB & 5.57m & 546.73MB  & 17.98\\
 & 1097146 & 7.45GB & 6.58m & 752.92MB & 17.73 \\
\bottomrule
\end{tabular}
\end{table}

\begin{figure*}[!t]
    \centering
    \includegraphics[width=\textwidth]{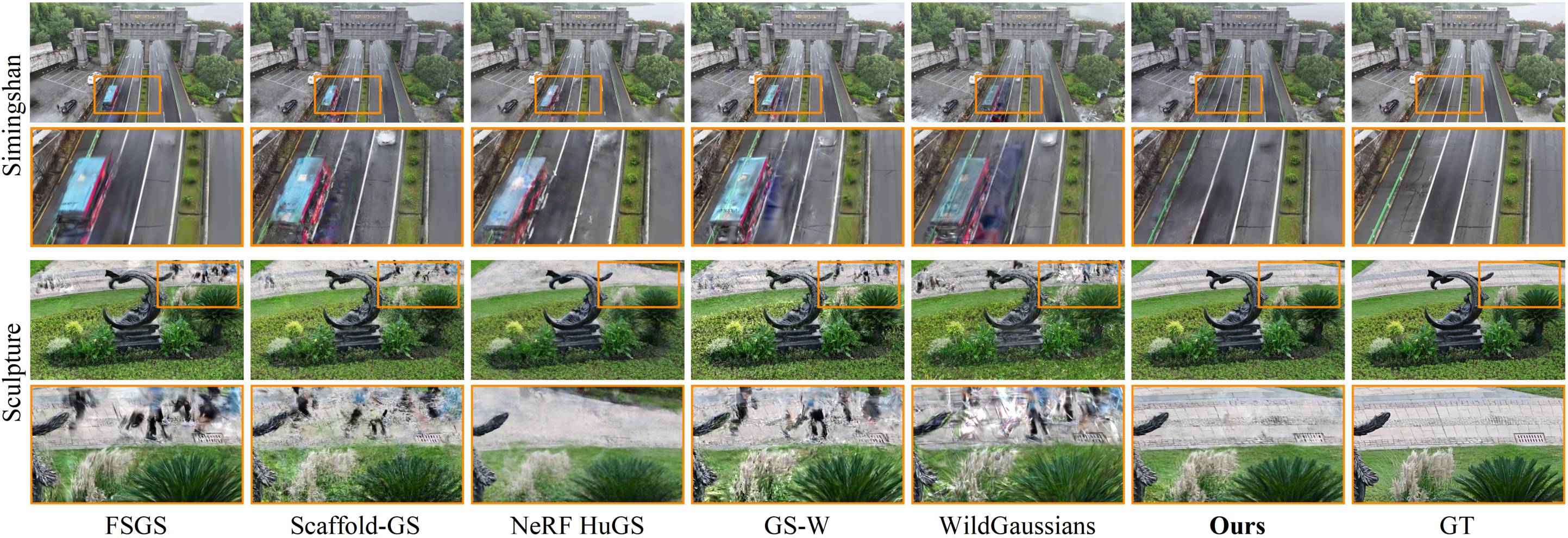}
    \caption{\textbf{Qualitative results on DronSplat dataset (dynamic) with limited views.} Each scene contains only six input views}
    \label{add_exp_fig}
\end{figure*}

\begin{table}[!h]
  \centering
  \renewcommand{\arraystretch}{1.3} 
  \caption{\textbf{Ablation study of training time for different modules across three GPU platforms.} We conduct the experiments on Simingshan and Sculpture of DroneSplat (dynamic) dataset. For comparison, WildGaussians requires 472 minutes for training, while GS-W takes 138 minutes.}
  \label{gpu_time}
  \resizebox{\linewidth}{!}{ 
  \begin{tabular}{
   cccc|ccc
  } 
    \toprule

                    Local  &  Global & Dense &  Voxel-guided &  \multicolumn{3}{c}{Traing Time}  \\ 
                     Masking &   Masking & Points &  Optimization & A100  & 4090 &  3090  \\
    \midrule
      \ding{55} &   \ding{55}  & \ding{55} & \ding{55} &  \large{7.02m}   &   \large{5.13m}  &  \large{9.63m}  \\
      \checkmark &  \ding{55}  & \ding{55}  & \ding{55}   &  \large{15.30m}   &   \large{14.01m}  &  \large{19.87m}   \\
      \checkmark &  \checkmark   &   \ding{55}  & \ding{55}  &  \large{18.25m}   &  \large{15.83m}   &  \large{25.93m}   \\
      \checkmark &  \checkmark   &   \checkmark  & \ding{55}  &  \large{16.50m}  &  \large{14.77m}   &  \large{21.70m}\\
      \checkmark &  \checkmark   &   \checkmark  & \checkmark  &  \large{16.62m}  &  \large{14.83m}   &  \large{22.18m}\\
    \bottomrule
  \end{tabular}
  }
\end{table}

\subsection{Performance on Two Challenges}
In the main paper,  we separate the two challenges of \textit{scene dynamics} and \textit{viewpoint sparsity}, and compare them with the most advanced methods in the corresponding fields. Our consideration is that DroneSplat is the only framework capable of tackling both challenges currently. However, to demonstrate the superiority and robustness of our approach in tackling these two challenges simultaneously, we select two representative scenes from the DroneSplat dataset, \textit{Simingshan} and \textit{Sculpture},  to conduct comparative studies on non-static scene reconstruction with limited views. Each scene includes only six input views.

We compare the best-performing baselines from the Distractor Elimination and Limited-View Reconstruction experiments for non-static scene reconstruction with limited views. As shown in Table \ref{add_exp} and Figure \ref{add_exp_fig}, our method outperforms the baselines in both scenes. 
NeRF-HuGS \cite{chen2024nerf}, GS-W \cite{zhang2024gaussian} and WildGaussians \cite{kulhanek2024wildgaussians} fail to effectively eliminate dynamic distractors under sparse-view conditions. 

\begin{table}[!ht]
  \centering
  \renewcommand{\arraystretch}{1.2} 
  \caption{\textbf{Quantitative results on DronSplat dataset (dynamic) with limited views.} The \colorbox{red!30}{1st}, \colorbox{orange!30}{2nd} and \colorbox{yellow!30}{3rd} best results are highlighted.}
  \resizebox{\linewidth}{!}{ 
  \begin{tabular}{
    l|c|c|c|c|c|c
  } 
    \toprule
                     \multirow{2}{*}{Method}  & \multicolumn{3}{c|}{\large{Simingshan}} & \multicolumn{3}{c}{\large{Sculpture}} \\
                     & PSNR$\uparrow$ & SSIM$\uparrow$ & LPIPS$\downarrow$ & PSNR$\uparrow$ & SSIM$\uparrow$ & LPIPS$\downarrow$   \\ 
    \midrule
    FSGS \small{[ECCV'24]}  &   \cellcolor{yellow!30}\large{19.97}   & \large{0.632}   &  \large{0.307}   &   \cellcolor{yellow!30}\large{16.55}  &  \cellcolor{orange!30}\large{0.373}    &  \cellcolor{yellow!30}\large{0.312} \\
    Scaffold-GS \small{[CVPR'24]}   &   \large{19.59}   &  \cellcolor{yellow!30}\large{0.637}  &  \cellcolor{orange!30}\large{0.197}   &  \large{16.04}   &  \cellcolor{yellow!30}\large{0.321}    &  \cellcolor{orange!30}\large{0.301} \\
    \midrule
    NeRF-HuGS \small{[CVPR'24]}  &   \cellcolor{orange!30}\large{20.41}   &  \cellcolor{orange!30}\large{0.656}  &  \cellcolor{yellow!30}\large{0.236}   &  \cellcolor{orange!30}\large{16.89}   &  \large{0.249}    &  \large{0.458}  \\
    GS-W \small{[ECCV'24]}   &   \large{19.21}   &  \large{0.602}  &  \large{0.292}   &  \large{14.84}   &  \large{0.269}    &  \large{0.329}  \\
    WildGaussians \small{[NIPS'24]} &    \large{18.91}  &  \large{0.582}  &   \large{0.286}  &   \large{15.58}  &   \large{0.255}   &  \large{0.409} \\
    Ours               &  \cellcolor{red!30}\large{22.46}   &  \cellcolor{red!30}\large{0.728}  &  \cellcolor{red!30}\large{0.156}   &   \cellcolor{red!30}\large{18.39}  &   \cellcolor{red!30}\large{0.427}   &  \cellcolor{red!30}\large{0.253} \\
    \bottomrule
  \end{tabular}
  }
  \label{add_exp}
\end{table}

\section{Limitations and Future Work}
Our framework leverages effective heuristics and the powerful capabilities of the segmentation model. However, it does have certain limitations in some cases. Firstly, in the video segmentation of Complement Global Masking, we track high-residual objects within their context. In practice, we find that tracking small objects is often unreliable, and tracking errors can sometimes lead to suboptimal results. A potential improvement could involve adding a post-processing to extract features of the tracked targets and filter out results with significant feature discrepancies.

In addition, our approach eliminates the influence of dynamic objects on static scene reconstruction by identifying and masking them. However, if a region is consistently occupied by dynamic distractors, it may lead to underfitting in the reconstruction. To address this, another possible improvement could be the integration of diffusion models to inpaint such regions.

\section{Additional Qualitative Results}
We additionally show the visualization results of our comparison experiment (Sec \ref{sec:compare}).
\subsection{DroneSplat Dataset (dynamic)}
\begin{figure*}[!h]
    \centering
    \includegraphics[width=\textwidth]{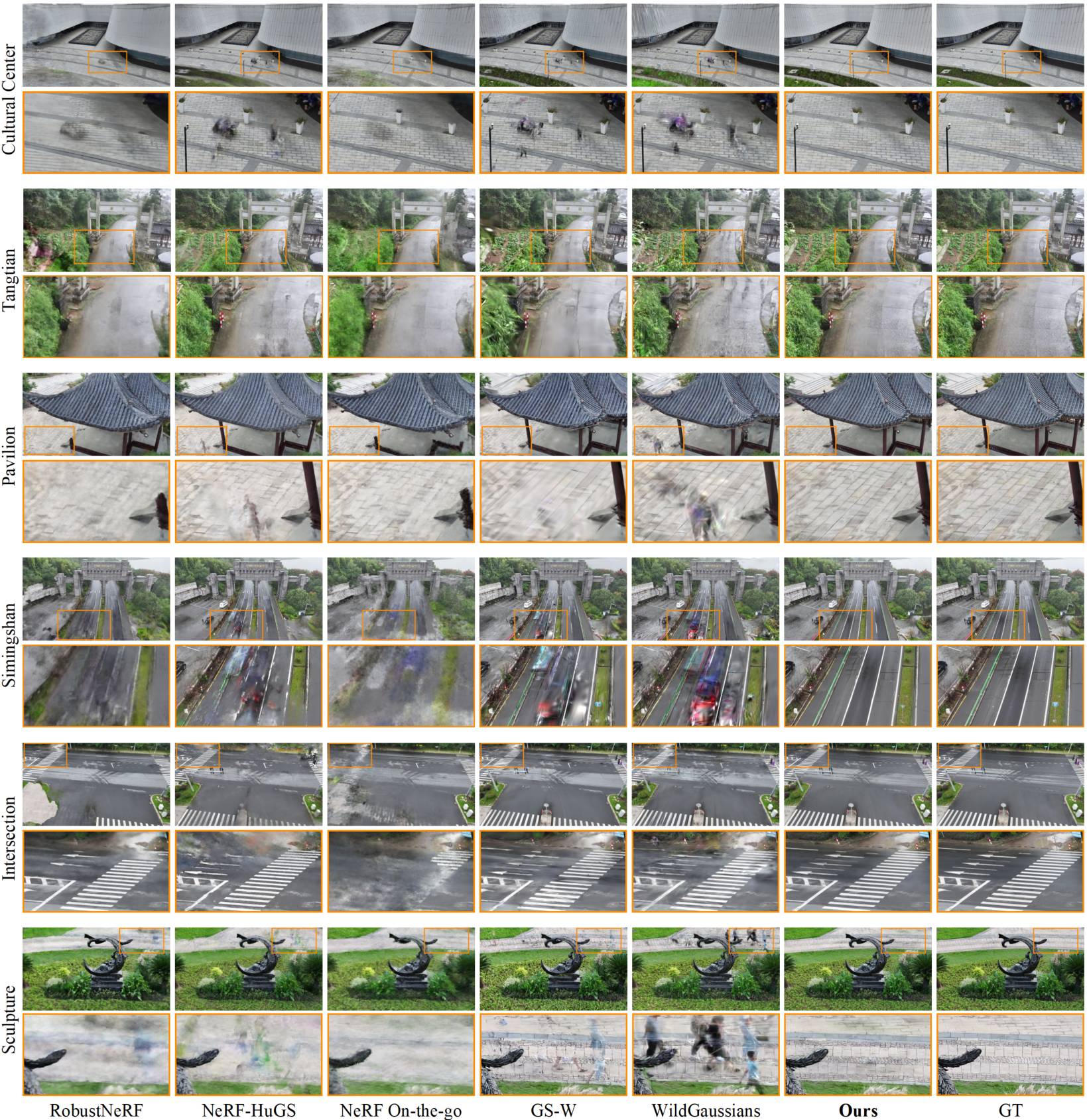}
    \caption{\textbf{Qualitative results of distractor elimination on DronSplat dataset (dynamic).} }
    \label{supp_comapre1}
\end{figure*}

As shown in Figure \ref{supp_comapre1}, NeRF-based methods, such as RobustNeRF \cite{sabour2023robustnerf} and NeRF On-the-go \cite{ren2024nerf}, eliminate dynamic objects in the scene but lack detail, sometimes even missing parts of the scene. A typical example is the \textit{pavilion} scene, where RobustNeRF and NeRF On-the-go successfully remove dynamic pedestrians but fail to reconstruct certain areas accurately, such as the missing the pillars of the pavilion. GS-W \cite{zhang2024gaussian} and WildGaussians \cite{kulhanek2024wildgaussians} perform poorly in the presence of numerous dynamic distractors, failing to eliminate them effectively. In contrast, our method not only effectively removes dynamic objects but also preserves fine scene details with high fidelity.

\subsection{NeRF On-the-go Dataset}
\begin{figure*}[!h]
    \centering
    \includegraphics[width=\textwidth]{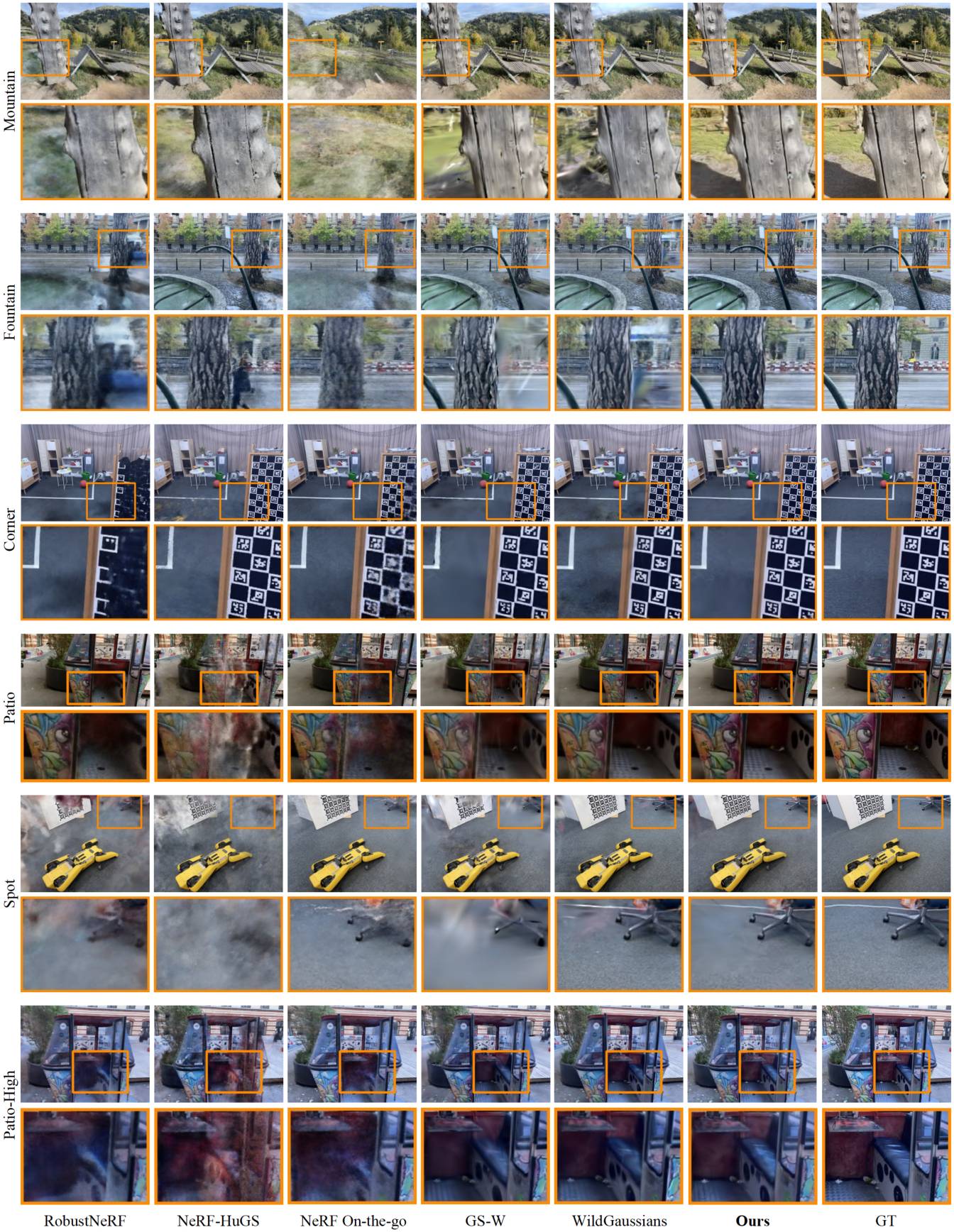}
    \caption{\textbf{Qualitative results of distractor elimination on NeRF On-the-go dataset.} }
    \label{supp_comapre2}
\end{figure*}

Compared to our DroneSplat dataset (dynamic), the NeRF On-the-go dataset \cite{ren2024nerf} features a greater number of viewpoints, with some scenes exhibiting higher levels of occlusion. As shown in the figure \ref{supp_comapre2}, NeRF-HuGS \cite{chen2024nerf} struggles to handle highly occluded scenes, resulting in blurring and artifacts. While WildGaussians \cite{kulhanek2024wildgaussians} effectively eliminates dynamic distractors and preserves scene details in most cases, it produces inaccurate reconstructions in areas with sparse viewpoints, such as the roadside in the \textit{fountain} scene. 
Our method demonstrates superior and robustness performance in handling varying levels of occlusion compared to existing approaches.

\subsection{DroneSplat Dataset (static)}
\begin{figure*}[!h]
    \centering
    \includegraphics[width=\textwidth]{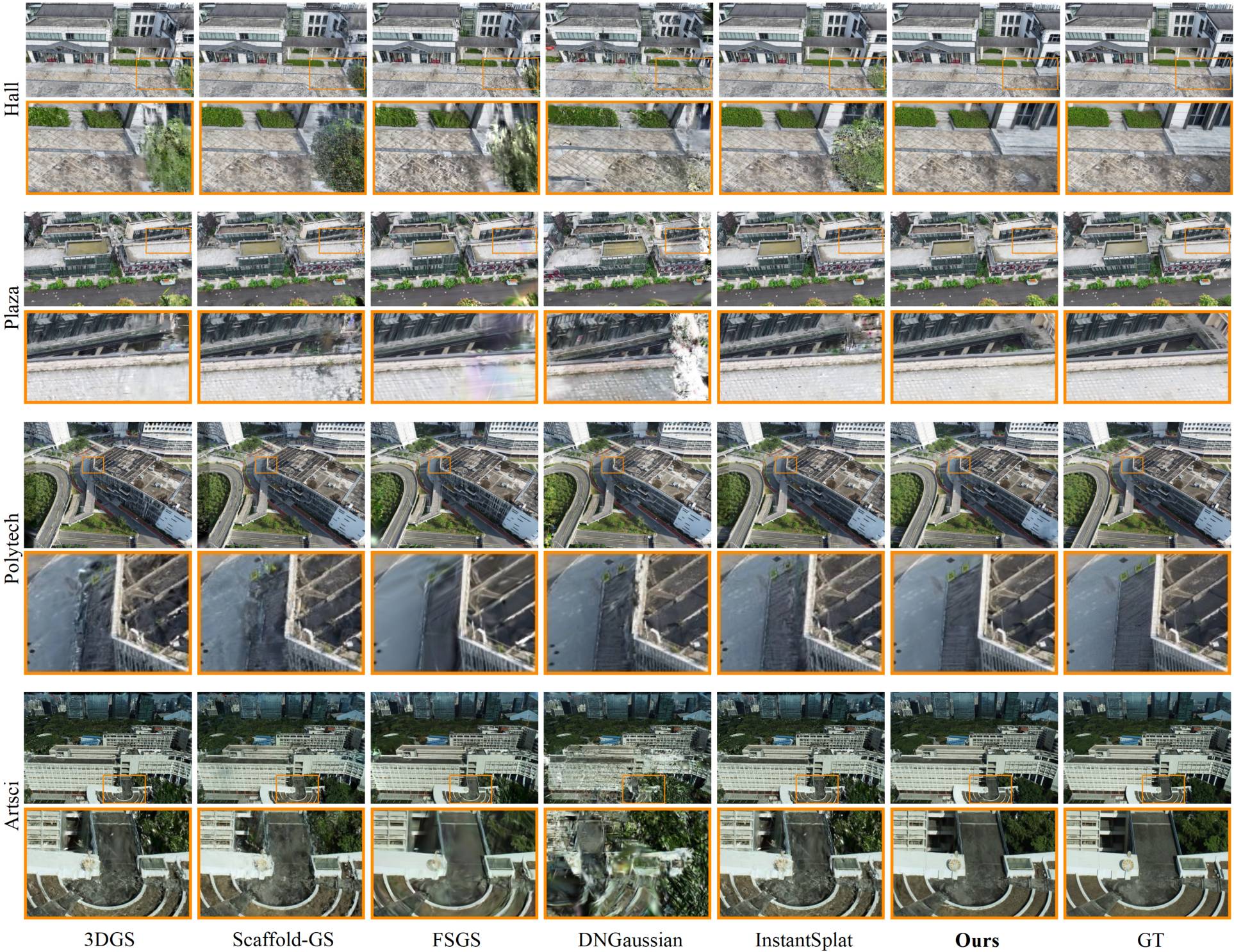}
    \caption{\textbf{Qualitative results of limited-view reconstruction on DronSplat dataset (static) and UrbanScene3D dataset.} }
    \label{supp_comapre3}
\end{figure*}

To simulate the challenges encountered in practical use, the experimental scenes we select have small overlap between viewpoints. As shown in the figure \ref{supp_comapre3}, leveraging rich geometric priors and the Voxel-guided Optimization strategy, our method reconstructs geometrically accurate scenes even under limited views constraint.

\subsection{UrbanScene3D Dataset (static)}
The results and conclusions are similar to those on the DroneSplat dataset (static).

\end{document}